\documentclass{article}

\PassOptionsToPackage{numbers,sort&compress}{natbib}
 \usepackage[main, final]{neurips_2026}
\usepackage[utf8]{inputenc} % allow utf-8 input
\usepackage[T1]{fontenc}    % use 8-bit T1 fonts
\usepackage{hyperref}       % hyperlinks
\usepackage{url}            % simple URL typesetting
\usepackage{booktabs}       % professional-quality tables
\usepackage{amsfonts}       % blackboard math symbols
\usepackage{nicefrac}       % compact symbols for 1/2, etc.
\usepackage{microtype}      % microtypography
\usepackage{xcolor}         % colors

\usepackage{graphicx}
\usepackage{booktabs}
\usepackage{stmaryrd}
\usepackage{color}
\usepackage{tabularray}
\usepackage{rotating}
\usepackage{subcaption}
\usepackage{hyperref}
\usepackage{amsmath}
\usepackage[dvipsnames]{xcolor}

\title{RoSplat: Robust Feed-Forward Pixel-wise Gaussian Splatting for Varying Input Views and High-Resolution Rendering}

\newcommand{\figimg}[1]{\raisebox{-0.5\height}{\includegraphics[width=\linewidth]{#1}}}
\newcommand{\rotlabel}[1]{\raisebox{-0.5\height}{\rotatebox[origin=c]{90}{#1}}}

\author{
\text{Hoang Chuong Nguyen}\textsuperscript{1} \hspace{0.5cm} \text{Renjie Wu}\textsuperscript{1} \hspace{0.5cm}  \text{Jose M. Alvarez}\textsuperscript{2}  \hspace{0.5cm}
\text{Miaomiao Liu}\textsuperscript{1}\\
\textsuperscript{1}\text{Australian National University} \hspace{1.0cm}  \textsuperscript{2}\text{NVIDIA}\\
{\tt\small hoangchuong.nguyen@anu.edu.au \hspace{0.1cm} renjie.wu@anu.edu.au } \\
{\tt\small josea@nvidia.com \hspace{0.1cm}  miaomiao.liu@anu.edu.au}   
}

\begin{document}

\maketitle

\begin{abstract}
Generalizable 3D Gaussian Splatting has recently emerged as an efficient approach for novel-view synthesis, enabling feed-forward synthesis from only a few input views. However, existing pixel-wise feed-forward methods suffer from over-bright renderings when the number of input views varies during inference, as well as insufficient supervision for accurate Gaussian scale estimation, which leads to hole artifacts, particularly in high-resolution renderings. To address these issues, we identify that the over-brightness is caused by the varying number of overlapping Gaussians and propose a simple alpha normalization strategy to maintain brightness consistency across different number of input views. In addition, we introduce an auxiliary 3D sampling-based regularizer to improve Gaussian scale estimation, thereby mitigating hole artifacts in high-resolution rendering. Experiments on benchmark datasets demonstrate that our method significantly improves baseline models under varying input-view and high-resolution rendering settings.

\end{abstract}

\section{Introduction}
\label{sec:intro}
% Why we need this tasks. 
% Application of NVS
Novel-view synthesis (NVS) is a foundational computer vision task that plays a crucial role in many modern applications, ranging from augmented reality to robotics and autonomous driving. 
% 3DGS, but rely on per-scene optimization, and struggle in sparse-view setting. 
3DGS~\cite{kerbl20233d} has emerged as a widely-used technique thanks to its real-time rendering speed and high-fidelity rendering results. Despite recent advances~\cite{yu2024mip,Huang2DGS2024,chao2025texturedgaussians}, these methods still rely on a per-scene optimization scheme and constrain their generalization ability in real applications. 

\begin{figure}[t]
    \centering
    \includegraphics[width=\textwidth]{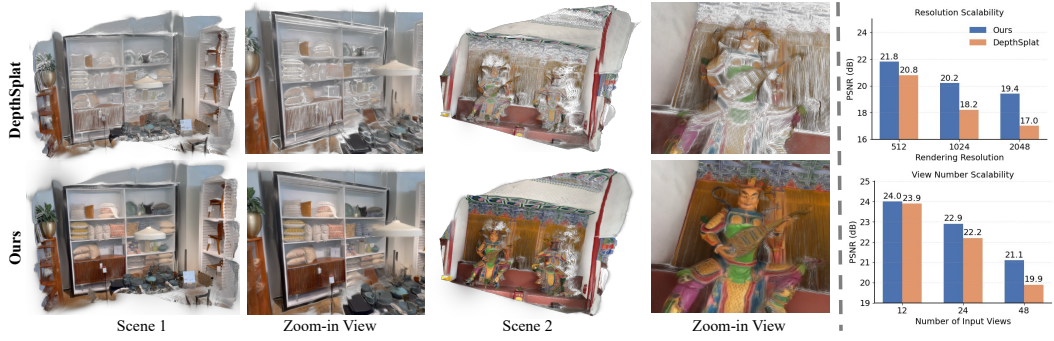}
    \caption{\textbf{Left}: DepthSplat~\cite{xu2025depthsplat} exhibits hole artifacts due to its predicted small-scale Gaussians. Our method mitigate the hole issues and produce more complete view. \textbf{Right}: Results overview: As the number of input views and rendering resolution increase, our method consistently achieves better image quality than DepthSplat.}
    \label{fig:teaser}
\end{figure}

% Why need generalizable 3DGS. 
Recently, several feed-forward Gaussian Splatting methods~\cite{charatan2024pixelsplat,chen2024mvsplat,xu2025depthsplat} have been introduced to bypass the reliance on the per-scene optimization process. These methods predict pixel-wise Gaussians from a few input images in a feed-forward manner. By leveraging a large scale training dataset, such methods learn strong priors that enable generalization across diverse test scenarios, particularly under sparse-view settings, where the model can reconstruct the scene appearance given as few as two input views. 
% Issues of recent generalizable 3DGS methods
Despite recent progress, existing methods~\cite{xu2025depthsplat, zhang2025transplat,chen2024mvsplat} still suffer from the following limitations: (1) over-bright rendering due to the limited generalization ability to varying numbers of input views during inference, and (2) hole artifacts in high-resolution rendering due to the insufficient supervision for accurately estimating Gaussian scales under sparse-view settings. Fig.~\ref{fig:teaser} shows an example where the existing work~\cite{xu2025depthsplat} produce holes when zooming into a particular scene region. 
 
% To address these issues, we first \ml{theoretically} analyze the over-brightness issue \ml{via the rendering process for pixel-wise feed-forward methods} when the number of input views varies during inference. \ml{Based on our analysis, pixel-wise feed-forward methods tend to generate more overlapping Gaussians for the same 3D region. It leads to a derived alpha normalization to reduce this issue.}
% %Pixel-wise feed-forward methods tend to generate more overlapping Gaussians for the same 3D region, which increases the accumulated compositing weights during rendering.  
% Existing approaches attempt to reduce such redundancy by fusing Gaussians in the overlapping regions using either graph neural network~\cite{zhang2024gaussian} or a pixel-alignment structure~\cite{wang2024freesplat}. While effective, these solutions rely on carefully designed network structures for \ml{reducing guassians}. In contrast, we introduce a simple yet \ml{theoretically grounded} alpha normalization strategy that adjusts each Gaussian’s contribution based on the number of overlapping Gaussians in 3D space. This preserves the total accumulated compositing weight across varying numbers of input views, thereby addressing the issue of over-brightness and improving the generalization. Our proposed normalization is considered as a soft-constraint during training and could be applied to the existing feed-forward pixel-wise Gaussian splatting methods.

To address these issues, we first provide a theoretical analysis of the over-brightness issue arising in the rendering process of pixel-wise feed-forward methods when the number of input views varies during inference. Based on our analysis, pixel-wise feed-forward methods tend to generate more overlapping Gaussians for the same 3D region, leading to increased accumulated compositing weights during rendering. This observation motivates our derived alpha normalization strategy to alleviate the issue. Existing approaches attempt to reduce such redundancy by fusing Gaussians in overlapping regions using either graph neural networks~\cite{zhang2024gaussian} or pixel-alignment structures~\cite{wang2024freesplat}. While effective, these solutions rely on carefully designed network architectures for reducing redundant Gaussians. In contrast, we introduce a simple yet theoretically grounded alpha normalization strategy that adjusts each Gaussian’s contribution according to the number of overlapping Gaussians in 3D space. This preserves the total accumulated compositing weight across varying numbers of input views, thereby mitigating over-brightness and improving generalization. Our normalization is formulated as a soft constraint during training and can be readily applied to existing feed-forward pixel-wise Gaussian splatting methods.

To mitigate the hole artifacts in rendering high-resolution images, we introduce an auxiliary 3D sampling-based regularizer for Gaussian scale estimation. Specifically, we augment the standard 3DGS~\cite{kerbl20233d} rendering process with a 3D sampling-based rendering branch that produces additional rendered images for computing a supplementary color rendering loss. 
Our goal is to encourage the predicted 3D Gaussians to enclose the sampled 3D points close to each surface point. This loss provides more direct supervision for Gaussian scales, encouraging sufficient spatial support for observed 3D regions, thereby reducing hole artifacts in rendered images. Note that this regularization branch is used only during training and can be discarded during inference, thereby preserving the efficiency of the standard 3DGS rendering process~\cite{kerbl20233d}. Both the proposed alpha normalization and 3D sampling-based regularizer are architecture-agnostic, and can be integrated with existing methods.

% Contribution
In summary, our contributions are: (i) we derive a theoretically grounded \textbf{alpha normalization} approach that reweights each Gaussian’s contribution based on the number of overlapping Gaussians, improving robustness to varying input-view counts during training and inference; and (ii) we introduce a \textbf{3D sampling-based regularizer}, which provides more direct supervision for Gaussian scales estimation and reduces hole artifacts by encouraging sufficient spatial support over the observed 3D scene regions. Experiments on benchmark datasets demonstrate that our method significantly improves baseline models when evaluated under varying input-view counts and high-resolution rendering settings.

\section{Related Work}

\subsection{Single-Scene Novel-View Synthesis} % Is this per-scene NVS?
% Many NVS methods reconstruct scenes through per-scene optimization from multi-view images. 
Neural Radiance Fields (NeRFs)~\cite{mildenhall2021nerf} 
have greatly advanced novel view synthesis and achieve high-fidelity rendering, they depend on dense input views and computationally intensive volume rendering, resulting in slow training and substantial memory consumption.
Several later NeRF variants~\cite{barron2021mip, reiser2021kilonerf, barron2023zip} were proposed to improve the efficiency and reconstruction performance.
Recently, Gaussian Splatting (GS)~\cite{kerbl20233d} and its variants~\cite{huang20242d, mallick2024taming, yu2024mip} introduce an explicit representation based on anisotropic Gaussians combined with differentiable rasterization-based rendering. 
However, both NeRFs and GS-based methods are inherently designed for dense-view supervision and require costly per-scene optimization. 
While some efforts~\cite{fan2024instantsplat, park2025dropgaussian, zhang2024cor} attempt to extend these approaches to sparse-view settings, many~\cite{kong2025generative, zheng2025nexusgs} still rely heavily on strong priors from large foundation models and demand substantial training time. Moreover, their performance typically degrades under sparse inputs, and the scene-specific optimization process limits scalability and practical deployment in real-world applications. In this work, we focus on Generalizable Gaussian Splatting for novel view synthesis and aim to improve the Gaussian scale estimation under the sparse-view setting.

% represent scenes as continuous volumetric radiance fields parameterized by MLPs, mapping 3D coordinates and viewing directions to color and density. 

\subsection{Generalizable Novel-View Synthesis}
To overcome the limitations of single-scene optimization methods, PixelNeRF~\cite{yu2021pixelnerf} pioneered the Generalizable NeRF by introducing an image-conditioned network that predicts radiance fields in a single forward pass. Later, many generalizable NeRF works~\cite{chen2021mvsnerf, xu2024murf, wang2022attention} were proposed to enhance cross-view consistency and reconstruction quality. Nevertheless, those methods still depend on per-pixel volume sampling during rendering or require considerable computational overhead.
% To enable real-time rendering, some approaches replace NeRF with Gaussian primitives, such as Splatter Image~\cite{szymanowicz2024splatter}, PixelSplat~\cite{charatan2024pixelsplat}, MVSplat~\cite{chen2024mvsplat}, and Depth-Splat~\cite{xu2025depthsplat} regress Gaussian parameters from single or sparse multi-view inputs.
To enable real-time rendering, a line of work replaces NeRF with Gaussian primitives. Representative methods, such as PixelSplat~\cite{charatan2024pixelsplat}, MVSplat~\cite{chen2024mvsplat}, Depth-Splat~\cite{xu2025depthsplat}, HiSplat~\cite{tang2024hisplat}, and TransSplat~\cite{zhang2025transplat}, directly regress pixel-aligned Gaussian from sparse multi-view inputs to enable feed-forward rendering without per-scene optimization. While the existing methods can accept a varying number of input views during inference, the rendered images show over-brightness issues due to multiple overlapping Gaussians contributing to the same 3D region. 
Recent works~\cite{wangzpressor, huang2025longsplat} attempt to mitigate issues of reducing redundant Gaussians through Gaussian merging or compression strategies. 
While effective, they require carefully designing a module to fuse overlapping Gaussians. In this paper, we propose alpha normalization and 3D sampling-based regularizer, which are architecture-agnostic, and can be integrated with existing feed-forward pixel-wise Gaussian Splatting methods to address the over-brightness issue and improve the Gaussian scale estimation to mitigate hole artifacts.
%However, they all overlook rendering artifacts that emerge as the number of input views increases, such as over-brightness caused by redundant splats and hole artifacts when rendering off-trajectory views. 

% Some studies~\cite{noposplat, huang2025no} also try to relax the requirement of known camera poses.
% However, they all overlook rendering artifacts that emerge as the number of input views increases, such as over-brightness caused by redundant splats and hole artifacts when rendering off-trajectory views. 
% In contrast, our work analyzes these failure modes and proposes solutions to address both redundancy-induced brightness accumulation and hole artifacts rendered in off-trajectory views.

%  zhang2024gaussian, tang2024hisplat, zhang2025transplat

% using pixel-aligned designs, epipolar transformers, or cost volumes. 

% In contrast, our work explicitly analyzes these failure modes under dense-view generalization and proposes solutions to address both redundancy-induced brightness accumulation and geometric incompleteness in off-trajectory view synthesis.

% \subsection{Robust and Generalizable Gaussian Splatting}

\section{Preliminary: Feed-forward Pixel-wise Gaussian Predictions}
\label{sec:preliminary}

Given $K$ input images $\{ \mathbf{I}^{i} \in \mathbb{R}^{H\times W \times 3} \}_{i=1}^K$ and their corresponding camera projection matrices $\{ \mathbf{P}^{i}=\mathbf{K}^{i}[\mathbf{R}^{i}|\mathbf{t}^i] \in \mathbb{R}^{3\times 4} \}_{i=1}^K$ (with $\mathbf{K}^i \in \mathbb{R}^{3\times 3}$ being a camera intrinsic matrix and $\mathbf{R}^i \in \mathbb{SO}(3)$, $\mathbf{t}^i \in \mathbb{R}^3$ denoting a camera rotation matrix, translation vector, respectively), a feed forward model $f_{\theta}$ is used to predict pixel-wise 3D Gaussians $\mathcal{G}^{\mathrm{3D}}_j$: 
\begin{equation}
    \{ \mathcal{G}^{3D}_j \}_{j=1}^{H \times W \times K} = \left\{(\boldsymbol{\mu}_j, o_j, \mathbf{s}_j, \mathbf{q}_j, \mathbf{c}_j)\right\}_{j=1}^{H \times W \times K} 
    = f_{\theta}\!\left(\left\{(\mathbf{I}_i,\mathbf{P}_i)\right\}_{i=1}^{K}\right)
\end{equation}
% where $\boldsymbol{\mu}_j\in \mathbb{R}^3$, $o_j\in\mathbb{R}$, $\mathbf{s}_i\in \mathbb{R}^3$, $\mathbf{q}_i\in\mathbb{R}^{4}$, $\mathbf{c}_j$ denotes the predicted Gaussians center, opacity, covariance matrix
where $\boldsymbol{\mu}_j$, $o_j$, $\mathbf{s}_j$, $\mathbf{q}_j$, $\mathbf{c}_j$ denotes the Gaussian's center, opacity, scales, rotations (represented via quaternion), and color (represented as spherical harmonics), respectively. Additionally, the Gaussian center $\boldsymbol{\mu}$ in view ${i}$ is obtained by back-projecting a pixel to 3D using a predicted depth map $\mathbf{D}^{i}\in \mathbb{R}^{H\times W}$. Given the predicted Gaussians, existing feed-forward methods follow the standard rendering pipeline of 3DGS that first splats the 3D Gaussians $\mathcal{G}^{3D}_j$ onto image plane to obtain 2D Gaussians $\mathcal{G}^{\mathrm{2D}}_j$. Then, the color of a pixel $\mathbf{p}$ can be computed as follows, 
\begin{equation}
    \mathbf{c}(\mathbf{p}) = \sum_{j=1}^{N} w_j(\mathbf{p}) \mathbf{c}_j = \sum_{j=1}^{N}  \alpha_j(\mathbf{p}) T_j(\mathbf{p}) \mathbf{c}_j, \quad \text{where }\alpha_j(\mathbf{p}) = o_j \mathcal{G}^{\mathrm{2D}}_j(\mathbf{p}), 
    \label{eq:rendering_equation_3dgs}
\end{equation}
with $T_j= \prod_{k=1}^{j-1}\bigl(1 - \alpha_k\bigr)$ being the accumulated transmittance and $N$ indicating the number of Gaussians splatted to compute the color of pixel $\mathbf{p}$.

\begin{figure}[t]
    \centering
    \includegraphics[width=\textwidth]{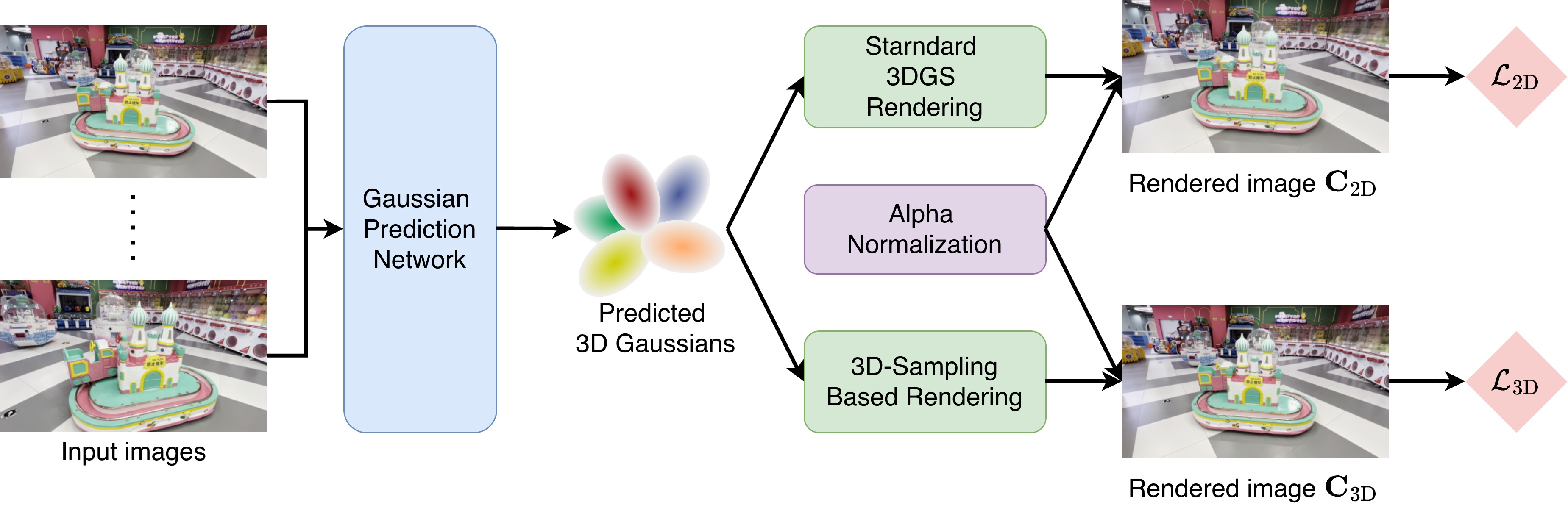}
    \caption{\textbf{Overall pipeline}. We introduce two components into the existing pixel-wise Gaussian prediction framework. First, \textbf{alpha normalization} is integrated into the rendering process to improve robustness to varying numbers of input views. Second, \textbf{a 3D sampling-based regularizer $\mathcal{L}_{\text{3D}}$} promotes accurate Gaussian scale estimation, mitigating hole artifacts under high-resolution rendering.}
    \label{fig:method_figure}
\end{figure}

\begin{figure}
    \centering
    \begin{minipage}[t]{0.5\textwidth}
        \centering
        \includegraphics[width=\linewidth]{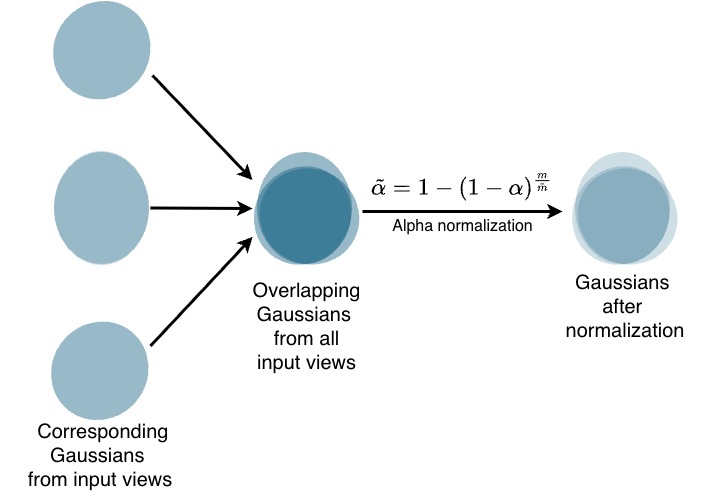}
        \caption{\textbf{Alpha normalization} adjusts each Gaussian's contribution based on its overlap count to avoid overbright rendering when increasing the number of input views.}
        \label{fig:method_alpha_norm}
    \end{minipage}
    \hfill
    \begin{minipage}[t]{0.42\textwidth}
        \centering
        \includegraphics[width=\linewidth]{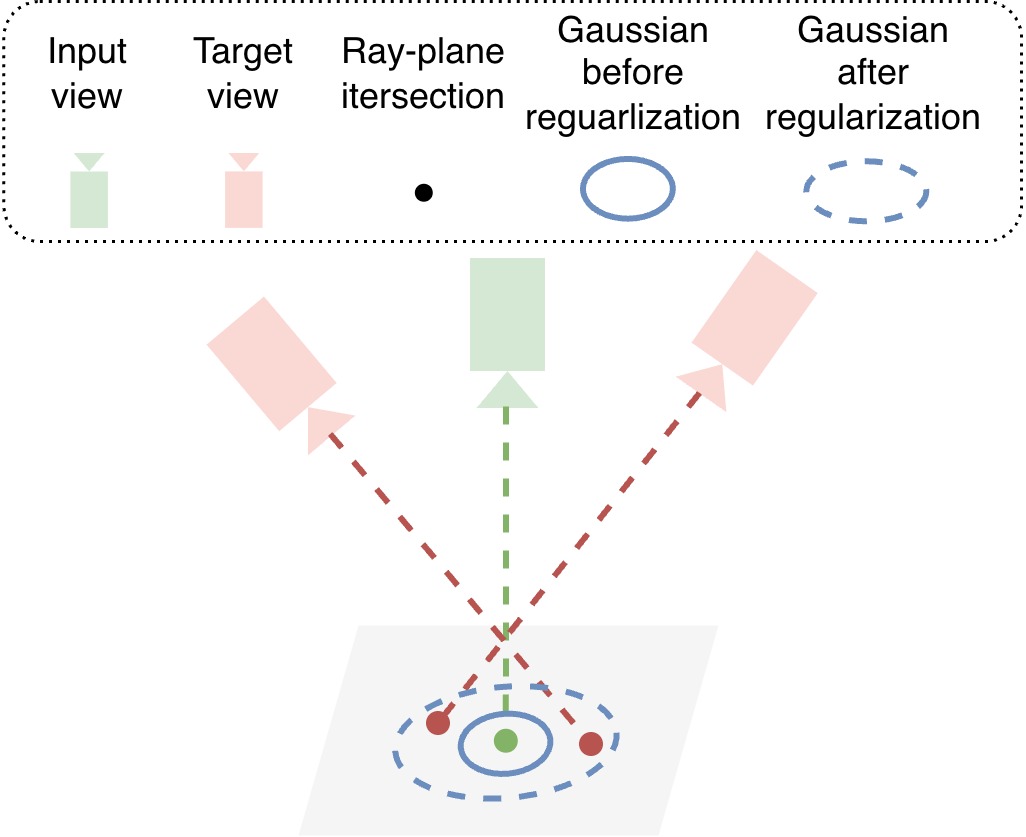}
        \caption{\textbf{3D sampling-based regularizer} encourages the model to predict sufficiently large scales of Gaussians to cover 3D points sampled on the same surface as the Gaussians.}
        \label{fig:method_3d_sampling}
    \end{minipage}
\end{figure}

\section{Method}
\label{sec:method}
% \subsection{Method Overview}
% will we talk
Built upon the feed-forward pixel-wise Gaussian prediction pipeline described in Sec.~\ref{sec:preliminary}, we propose two complementary components to improve the robustness of the feed-forward network with respect to varying numbers of input views and high-resolution rendering. The first component is a simple yet effective \textbf{alpha normalization} to alleviate the over-brightness issue. This proposal builds upon an analysis of the over-bright rendering when the test-time input-view count exceeds that during training. The second component is a \textbf{3D sampling-based regularizer} to address the hole artifacts, especially in high-resolution renderings. This regularization is implemented via an auxiliary 3D sampling-based rendering branch and an additional color rendering loss. 
Fig. ~\ref{fig:method_figure} depicts our overall framework, whereas Fig.~\ref{fig:method_alpha_norm} and Fig.~\ref{fig:method_3d_sampling} illustrate the alpha normalization and 3D sampling-based regularizer, respectively.

% \vspace{-4mm}
\subsection{Alpha Normalization}
\label{sub_sec:alpha_normnalization}
% How do you define number of overlapping Gaussians? 

Existing feed-forward pixel-wise 3D Gaussian Splatting methods~\cite{chen2021mvsnerf, zhang2025transplat, xu2025depthsplat} often struggle to generalize to varying numbers of input views at test time. Particularly, due to the pixel-wise Gaussian representation, increasing the number of input views typically produces more Gaussians that overlap on the same underlying 3D region, causing more Gaussians to contribute along the same pixel ray. As the sum of compositing weights $W=\sum_{j=1}^{N} w_j=1-\prod_{j=1}^{N}(1-\alpha_j)$ is a monotonically non-decreasing function of the number of Gaussians $N$, having more overlapping Gaussians tend to result in a higher accumulated compositing weight $W$. Moreover, since these overlapping Gaussians correspond to the same underlying 3D region, they often share similar color $\mathbf{c}_j(\mathbf{p})\approx\mathbf{\bar{c}}(\mathbf{p})$, therefore the rendered pixel color in Eq.~\ref{eq:rendering_equation_3dgs} can be approximated as $\mathbf{c}(\mathbf{p}) \approx \mathbf{\bar{c}(\mathbf{p})}W$. Under this approximation, the rendered color $\mathbf{c}(\mathbf{p})$ becomes a monotonically non-decreasing function of $W$, and thus a monotonically non-decreasing function of the number of overlapping Gaussians. Consequently, increasing the number of input views often raises the accumulated compositing weight $W$, which results in over-bright renderings and degraded image quality.

To alleviate this issue, we aim to keep the accumulated compositing weight $W$ approximately invariant when the number of overlapping Gaussians changes. As overlapping Gaussians represent the same 3D region, they tend to share similar geometric properties and are therefore approximately duplicates of each other. Under the ideal case where the overlapping Gaussians are exact duplicates of each other (i.e., $\alpha_j=\alpha$), the accumulated weight satisfies $W=1-\prod_{j=1}^{m}(1-\alpha_j)=1-(1-\alpha)^{m}$, where $m$ is the number of duplicate Gaussians. Assume the duplicate Gaussians count changes from $m$ to \(\tilde m\) due to changes in the number of input views. Under this setting, the accumulated weight becomes: 
\begin{equation}
    \tilde W=1-(1-\tilde\alpha)^{\tilde m}
\end{equation}

To ensure brightness invariance when the number of overlapping Gaussians varies, the following condition must be satisfied:
\begin{equation}
    1-(1-\tilde\alpha)^{\tilde m}=1-(1-\alpha)^m,
\end{equation}
leading to
\begin{equation}
    \tilde{\alpha}_j=1-(1-\alpha_j)^{\frac{m}{\tilde{m}_j}}
    \label{eq:alphaNormalisation}
\end{equation}
where $m$ is a hyperparameter indicating the number of overlapping Gaussians corresponding to the target accumulated compositing weight to be preserved (e.g., $m=2$ when a model is trained with two input views) and $\tilde{m}_j$ denotes the per-Gaussian overlap counts given the current set of input views. We refer to Eq.~\ref{eq:alphaNormalisation} as an alpha normalization approach to avoid the over-brightness issue.
%adjusts each Gaussian's contribution based on the number of overlapping Gaussian, so as to preserve the accumulated compositing weight and improve the model's robustness to varying input-view counts. 
% Specifically, the $\alpha_j$ value of each Gaussian used in the rendering process (see Eq.~\ref{eq:rendering_equation_3dgs}) are normalized as follows,
% \begin{equation}
%     \tilde{\alpha}_j=1-(1-\alpha_j)^{\frac{m}{\tilde{m}_j}}
% \end{equation}
% where $m$ is a hyper-parameter indicating the number of overlapping Gaussians corresponding to the target accumulated compositing weight we aim to preserve (e.g., $m=2$ if a model is trained with two input views) and $\tilde{m}_j$ denotes the per-Gaussian overlap counts given the set of input views. 

% Our alpha normalization approach is built upon the following reasoning. As overlapping Gaussians represent the same 3D region, they tend to share similar geometric properties and are therefore approximately duplicates of each other. Under an idealized case where the overlapping Gaussians are exact duplicates of each other, the accumulated compositing weight satisfies $W=1-\prod_{j=1}^{m}(1-\alpha)=1-(1-\alpha)^{m}$. In this case, alpha-normalization preserves the accumulated weight $W$ staying the same, when the number of duplicate Gaussians changes from $m$ to \(\tilde m\): 
% \begin{equation}
%     \tilde W=1-(1-\tilde\alpha)^{\tilde m}=1-\Big((1-\alpha)^{m/\tilde m}\Big)^{\tilde m}=1-(1-\alpha)^m=W.
% \end{equation}

To compute per-Gaussian overlap counts $\tilde{m}_j$, we rely on a depth-consistency check across input views. In particular, for each $i$-th input view, we compute a Gaussian count map $\mathbf{\tilde{M}}^{i}\in\mathbb{R}^{H\times W}$ as
\begin{equation}
    \mathbf{\tilde{M}}^i(\mathbf{p}) = \sum_{k=1}^{K} \left\llbracket\, 
    \frac{\left|\mathbf{D}^{i}(\mathbf{p}) - \mathbf{D}^{k\rightarrow i}(\mathbf{\hat{p}})\right|}
    {\mathbf{D}^{i}(\mathbf{p}) + \mathbf{D}^{k \rightarrow i}(\mathbf{\hat{p}})} 
    \leq \tau \,\right\rrbracket,
\end{equation}
where $\mathbf{\hat{p}}$ is a pixel in the $k^{\text{th}}$ input image corresponding to pixel $\mathbf{p}$, $\mathbf{D}^{k\rightarrow i}$ denotes the depth map of view $k$ expressed in the coordinate system of view $i$, $\llbracket\cdot\rrbracket$ is the Iverson bracket, and $\tau$ is a depth error threshold. Subsequently, the per-Gaussian overlap counts $\{\tilde{m}_j\}_{j=1}^{H\times W \times K}$ can be obtained from the count maps $\{\mathbf{\tilde{M}}^i\}_{i=1}^K$.

\subsection{3D Sampling-Based Regularizer}
\label{sub_sec:3D_sampling_based_regularization}

Apart from robustness to varying input-view counts, existing methods~\cite{yu2021pixelnerf, chen2021mvsnerf, zhang2025transplat, xu2025depthsplat} also suffer from insufficient supervision for accurate Gaussian scale estimation. Specifically, prior works tend to predict small-scale Gaussians that can explain target training viewpoints. However, when rendering at high resolution, these methods often produce holes in the rendered images because the predicted Gaussians provide insufficient spatial support for complete coverage of the scene regions visible in the input views. 
%Moreover, the learned Gaussians tend to 
%This issue can be attributed to the fact that existing generalizable methods are typically trained to minimize the color rendering loss on video data with smooth camera trajectories, which provides limited viewpoint diversity for learning accurate Gaussian scales.

% We propose to leverage not only the predicted depth for computing the Gaussian center, but also the corresponding surface normals to encourage the predicted Gaussians to have sufficiently large scales to cover scene region observed in the input views. Specifically, we propose to render an additional image by leveraging 3D points sampled on the local plane of each pixel, defined by the predicted depth and its associated surface normal. To obtain these 3D samples on a local surface plane, we intersect the plane with camera rays corresponding to neighboring pixels in the target view. Each sampled 3D point is computed as the intersection between local plane defined at a pixel in the input view, and the camera rays in a small neighborhood corresponding to the pixel in the target view. The ultimate goal is to encourage the 3D Gaussians of sufficient scale to enclose the sampled 3D point and contribute to the rendering of its color. In particular, our method augments the standard 3DGS rendering equation~\cite{kerbl20233d} by leveraging sampled 3D points as follows,

To mitigate this issue, we propose to leverage not only the predicted depth for computing the Gaussian center, but also the corresponding surface normals to encourage the predicted Gaussians to have sufficiently large scales to cover scene region observed in the input views. Specifically, we propose to render an additional image by leveraging 3D points sampled on the local plane of each pixel, defined by the predicted depth and its associated surface normal. These sampled 3D points are computed as intersections between the local plane defined at a pixel in the input view, and the camera rays passing through a small neighborhood corresponding to that pixel in the target view. The ultimate goal is to encourage the 3D Gaussians of sufficient scale to enclose the sampled 3D points and contribute to the rendering. In particular, our method augments the standard 3DGS rendering equation~\cite{kerbl20233d} by leveraging sampled 3D points as follows,
\begin{equation}
    \mathbf{{c}}^{\text{3D}}(\mathbf{p}) = \sum_{j=1}^{N}  \alpha^{\text{3D}}_j(\mathbf{x}_j(\mathbf{p}))\, T_j^{\text{3D}}(\mathbf{x}_j(\mathbf{p}))\, \mathbf{c}_j,
\end{equation}
\begin{equation}
    \text{with } \quad \alpha^{\text{3D}}_j(\mathbf{x}_j(\mathbf{p})) = o_j^{\text{3D}}\, \mathcal{G}^{\mathrm{3D}}_j(\mathbf{x}_j(\mathbf{p})), 
\end{equation}
\begin{equation}
    \quad \mathbf{x}_j(\mathbf{p})=\mathbf{o} + \frac{\mathbf{n}_j^T(\boldsymbol{\mu}_j-\mathbf{o})}{\mathbf{n}_j^T\mathbf{d}(\mathbf{p})}\mathbf{d}(\mathbf{p}),
\end{equation}
where $T_j^{\text{3D}}= \prod_{k=1}^{j-1}\bigl(1 - \alpha^{\text{3D}}_k\bigr)$, $\mathbf{x}_j(\mathbf{p})$ is the intersection between (i) a camera ray with direction $\mathbf{d}(\mathbf{p})$ originating from the camera center $\mathbf{o}$ and (ii) the plane defined by the Gaussian center $\boldsymbol{\mu}_j$ and a surface normal $\mathbf{n}_j$. To obtain $\mathbf{n}_j$, we first compute the normal map $\mathbf{N}^i \in \mathbb{R}^{H \times W \times 3}$ for each input view $i$ from the predicted depth map $\mathbf{D}^i$, and then assign $\mathbf{N}^i(\mathbf{p})$ as the normal of the Gaussian predicted at the same pixel location $\mathbf{p}$ in view $i$. We also introduce $o_j^{\text{3D}}$ as an additional Gaussian parameter predicted by the model to compensate for the discrepancy in magnitude between $\mathcal{G}^{2D}(\cdot)$ and $\mathcal{G}^{3D}(\cdot)$. 

% We then obtain the rendered image  $\mathbf{\hat{C}}^{\text{3D}}$ by stacking all $\mathbf{\hat{c}}^{\text{3D}}(\mathbf{p})$. To avoid over-extending Gaussian scales in regions near object boundaries with abrupt depth changes, we further modulate $\mathbf{\hat{C}}^{\text{3D}}$ as follows,
% \begin{equation}
%     \mathbf{C}^{\text{3D}}(\mathbf{p})= \mathbf{O}(\mathbf{p})\mathbf{\hat{C}}^{\text{3D}}(\mathbf{p}) + (1-\mathbf{O}(\mathbf{p}))\mathbf{C^{\text{2D}}}(\mathbf{p})
% \end{equation}
% \begin{equation}
%     \mathbf{O}(\mathbf{p})= \left\llbracket\, 
%     \text{Var}(\mathbf{D}^{\text{target}},\mathbf{p}) \leq \tau_d \quad \land \quad \text{Var}(\mathbf{C}^{\text{gt}},\mathbf{p}) \leq \tau_c
%     \,\right\rrbracket,
% \end{equation}
% where $\mathbf{C}^{\text{gt}}$, $\mathbf{D}^{\text{target}}$ are the ground-truth image and the predicted depth map of the target viewpoint, respectively. $\text{Var}(\mathbf{X}, \mathbf{p})$ indicates the variance value computed within the $3\times 3$ patch in $\mathbf{X}$ centered at pixel $\mathbf{p}$. $\tau_d$ and $\tau_c$ are hyperparameters defining depth and color variance thresholds.  $\mathbf{C}^{\text{2D}}$ can be obtained by stacking all the rendered color $\mathbf{c}(\mathbf{p})$ obtained in Eq.~\ref{eq:rendering_equation_3dgs}. 

We then obtain the rendered image  $\mathbf{{C}}^{\text{3D}}$ by stacking all $\mathbf{{c}}^{\text{3D}}(\mathbf{p})$, which is then used to compute the following loss to regularize the Gaussian scales estimated by the model,
\begin{equation}
\mathcal{L}_{\text{3D}}=\text{MSE}(\mathbf{C}^{\text{3D}}, \mathbf{C}^{\text{gt}}) + \beta\, \text{LPIPS}(\mathbf{C}^{\text{3D}}, \mathbf{C}^{\text{gt }}).
\end{equation}
By constructing $\mathbf{C}^{\text{3D}}$ using the 3D Gaussian scales directly, rather than their projected counterparts, minimizing $\mathcal{L}_{\text{3D}}$ provides a more direct supervision for accurate 3D scale estimation. As a result, the model is encouraged to predict Gaussians with sufficiently large scales to ensure sufficient coverage of the 3D scene, thereby alleviating hole artifacts during high-resolution rendering.

% for learning accurate Gaussian scales. This encourages accurate scale predictions that ensure sufficient coverage of the 3D scene, thereby alleviating hole artifacts in high-resolution rendering.

% It's worth mentioning that the gradient from $\mathcal{L}_{\text{3D}}$ is only back-propagated to the predicted Gaussians scale $\mathbf{s}$ and the opacity $o^{\text{3D}}$. 

% Our method

% Why our method works

% \usepackage{color}
% \usepackage{tabularray}
\begin{table}[t]
\centering
\scriptsize
\caption{\textbf{Comparison on RealEstate10k dataset under varying number of input views}. During training, all models take~{\bf two} views as input. $\alpha-X$ denotes applying alpha normalization to the $X$ method at inference time without retraining. Alpha normalization shows consistent improvements. } 
\resizebox{\textwidth}{!}{
\begin{tblr}{
  colsep  = 1.5pt,
  rowsep  = 0.5pt,
  column{even} = {r},
  column{3} = {r},
  column{7} = {r},
  column{9} = {r},
  column{11} = {r},
  column{13} = {r},
  cell{1}{2} = {c=3}{c},
  cell{1}{5} = {c},
  cell{1}{6} = {c=3}{c},
  cell{1}{9} = {c=3}{c},
  cell{1}{12} = {c=3}{c},
  cell{2}{2} = {c},
  cell{2}{3} = {c},
  cell{2}{4} = {c},
  cell{2}{5} = {c},
  cell{2}{6} = {c},
  cell{2}{7} = {c},
  cell{2}{8} = {c},
  cell{2}{9} = {c},
  cell{2}{10} = {c},
  cell{2}{11} = {c},
  cell{2}{12} = {c},
  cell{2}{13} = {c},
  cell{2}{14} = {c},
  cell{3}{2} = {font=\bfseries},
  cell{3}{3} = {font=\bfseries},
  cell{3}{4} = {font=\bfseries},
  cell{3}{5} = {c},
  cell{4}{2} = {font=\bfseries},
  cell{4}{3} = {font=\bfseries},
  cell{4}{4} = {font=\bfseries},
  cell{4}{5} = {c},
  cell{4}{6} = {font=\bfseries},
  cell{4}{7} = {font=\bfseries},
  cell{4}{8} = {font=\bfseries},
  cell{4}{9} = {font=\bfseries},
  cell{4}{10} = {font=\bfseries},
  cell{4}{11} = {font=\bfseries},
  cell{4}{12} = {font=\bfseries},
  cell{4}{13} = {font=\bfseries},
  cell{4}{14} = {font=\bfseries},
  cell{5}{2} = {font=\bfseries},
  cell{5}{3} = {font=\bfseries},
  cell{5}{4} = {font=\bfseries},
  cell{5}{5} = {c},
  cell{6}{2} = {font=\bfseries},
  cell{6}{3} = {font=\bfseries},
  cell{6}{4} = {font=\bfseries},
  cell{6}{6} = {font=\bfseries},
  cell{6}{7} = {font=\bfseries},
  cell{6}{8} = {font=\bfseries},
  cell{6}{9} = {font=\bfseries},
  cell{6}{10} = {font=\bfseries},
  cell{6}{11} = {font=\bfseries},
  cell{6}{12} = {font=\bfseries},
  cell{6}{13} = {font=\bfseries},
  cell{6}{14} = {font=\bfseries},
  cell{7}{2} = {font=\bfseries},
  cell{7}{3} = {font=\bfseries},
  cell{7}{4} = {font=\bfseries},
  cell{8}{2} = {font=\bfseries},
  cell{8}{3} = {font=\bfseries},
  cell{8}{4} = {font=\bfseries},
  cell{8}{6} = {font=\bfseries},
  cell{8}{7} = {font=\bfseries},
  cell{8}{8} = {font=\bfseries},
  cell{8}{9} = {font=\bfseries},
  cell{8}{10} = {font=\bfseries},
  cell{8}{11} = {font=\bfseries},
  cell{8}{12} = {font=\bfseries},
  cell{8}{13} = {font=\bfseries},
  cell{8}{14} = {font=\bfseries},
  vline{2,6} = {1}{Black},
  vline{3,7,10,13} = {1}{},
  vline{2,5-6,9,12} = {1-9}{Black},
  hline{3,5,7,9} = {1,4,6,8,11,14}{},
  hline{3,5,7,9} = {2-3,7,9-10,12-13}{Black},
}
                                & 2 input views &       &       &            & 4~ input views &       &       & 8~~ input views &       &       & 16~~ input views &       &       \\
~                                & PSNR $\uparrow$          & SSIM $\uparrow$  & LPIPS $\downarrow$ & ~          & PSNR $\uparrow$           & SSIM $\uparrow$  & LPIPS $\downarrow$ & PSNR $\uparrow$            & SSIM $\uparrow$  & LPIPS $\downarrow$ & PSNR $\uparrow$             & SSIM $\uparrow$  & LPIPS $\downarrow$ \\
MVSplat~\cite{chen2021mvsnerf}                          & 26.39         & 0.869 & 0.128 & ~          & 23.38          & 0.855 & 0.156 & 21.16           & 0.818 & 0.191 & 19.76            & 0.780  & 0.222 \\
$\alpha$-MVSplat    & \SetCell{c} --         & \SetCell{c} -- & \SetCell{c} -- & \textbf{~} & 26.41          & 0.881 & 0.135 & 26.06           & 0.879 & 0.145 & 24.83            & 0.859 & 0.168 \\
TranSplat~\cite{zhang2025transplat}                        & 26.69         & 0.875 & 0.125 & ~          & 23.74          & 0.875 & 0.135 & 20.52           & 0.82  & 0.186 & 18.85            & 0.775 & 0.225 \\
$\alpha$-TranSplat  & \SetCell{c} --         & \SetCell{c} -- & \SetCell{c} -- & \textbf{~} & 27.79          & 0.906 & 0.109 & 27.45           & 0.908 & 0.114 & 26.03            & 0.891 & 0.135 \\
DepthSplat~\cite{xu2025depthsplat}                      & 27.34         & 0.887 & 0.115 &            & 25.53          & 0.892 & 0.121 & 22.85           & 0.849 & 0.156 & 20.97            & 0.788 & 0.199 \\
$\alpha$-DepthSplat & \SetCell{c} --         & \SetCell{c} -- & \SetCell{c} -- & \textbf{~} & 28.35          & 0.914 & 0.105 & 27.61           & 0.909 & 0.114 & 25.86            & 0.875 & 0.150 \\
GGN~\cite{zhang2024gaussian}                              & 24.07         & 0.818 & 0.178 &            & 23.99          & 0.817 & 0.194 & 22.88           & 0.794 & 0.215 & 21.36            & 0.763 & 0.236 
\end{tblr}
}
\label{tab:result_alpha_norm_noPretrained_re10k}
\end{table}

\begin{table}[t]
\centering
\scriptsize
\caption{\textbf{Comparision on DL3DV dataset under varying number of input views}. During training, all model takes from 2 to 6 views as input. Our methods show consistent improvement over the baselines when increasing the number of input views. } 
\resizebox{\textwidth}{!}{
\begin{tblr}{
  colsep  = 1.5pt,
  rowsep  = 0.5pt,
  column{even} = {r},
  column{3} = {r},
  column{7} = {r},
  column{9} = {r},
  column{11} = {r},
  column{13} = {r},
  cell{1}{2} = {c=3}{c},
  cell{1}{5} = {c},
  cell{1}{6} = {c=3}{c},
  cell{1}{9} = {c=3}{c},
  cell{1}{12} = {c=3}{c},
  cell{2}{2} = {c},
  cell{2}{3} = {c},
  cell{2}{4} = {c},
  cell{2}{5} = {c},
  cell{2}{6} = {c},
  cell{2}{7} = {c},
  cell{2}{8} = {c},
  cell{2}{9} = {c},
  cell{2}{10} = {c},
  cell{2}{11} = {c},
  cell{2}{12} = {c},
  cell{2}{13} = {c},
  cell{2}{14} = {c},
  % cell{3}{2} = {font=\bfseries},
  % cell{3}{3} = {font=\bfseries},
  % cell{3}{4} = {font=\bfseries},
  cell{3}{5} = {c},
  % cell{4}{2} = {font=\bfseries},
  % cell{4}{3} = {font=\bfseries},
  % cell{4}{4} = {font=\bfseries},
  cell{4}{5} = {c},
  % cell{4}{6} = {font=\bfseries},
  % cell{4}{7} = {font=\bfseries},
  % cell{4}{8} = {font=\bfseries},
  % cell{4}{9} = {font=\bfseries},
  % cell{4}{10} = {font=\bfseries},
  % cell{4}{11} = {font=\bfseries},
  % cell{4}{12} = {font=\bfseries},
  % cell{4}{13} = {font=\bfseries},
  % cell{4}{14} = {font=\bfseries},
  % cell{5}{2} = {font=\bfseries},
  % cell{5}{3} = {font=\bfseries},
  % cell{5}{4} = {font=\bfseries},
  cell{5}{5} = {c},
  % cell{6}{2} = {font=\bfseries},
  % cell{6}{3} = {font=\bfseries},
  % cell{6}{4} = {font=\bfseries},
  % cell{6}{6} = {font=\bfseries},
  % cell{6}{7} = {font=\bfseries},
  % cell{6}{8} = {font=\bfseries},
  % cell{6}{9} = {font=\bfseries},
  % cell{6}{10} = {font=\bfseries},
  % cell{6}{11} = {font=\bfseries},
  % cell{6}{12} = {font=\bfseries},
  % cell{6}{13} = {font=\bfseries},
  % cell{6}{14} = {font=\bfseries},
  % cell{7}{2} = {font=\bfseries},
  % cell{7}{3} = {font=\bfseries},
  % cell{7}{4} = {font=\bfseries},
  % cell{8}{2} = {font=\bfseries},
  % cell{8}{3} = {font=\bfseries},
  % cell{8}{4} = {font=\bfseries},
  % cell{8}{6} = {font=\bfseries},
  % cell{8}{7} = {font=\bfseries},
  % cell{8}{8} = {font=\bfseries},
  % cell{8}{9} = {font=\bfseries},
  % cell{8}{10} = {font=\bfseries},
  % cell{8}{11} = {font=\bfseries},
  % cell{8}{12} = {font=\bfseries},
  % cell{8}{13} = {font=\bfseries},
  % cell{8}{14} = {font=\bfseries},
  vline{2,6} = {1}{Black},
  vline{3,7,10,13} = {1}{},
  vline{2,5-6,9,12} = {1-9}{Black},
  hline{3,5,7} = {1,4,6,8,11,14}{},
  hline{3,5,7} = {2-3,7,9-10,12-13}{Black},
}
                                & 6 input views &       &       &            & 12~ input views &       &       & 24~~ input views &       &       & 48~~ input views &       &       \\
~                                & PSNR $\uparrow$          & SSIM $\uparrow$  & LPIPS $\downarrow$ & ~          & PSNR $\uparrow$           & SSIM $\uparrow$  & LPIPS $\downarrow$ & PSNR $\uparrow$            & SSIM $\uparrow$  & LPIPS $\downarrow$ & PSNR $\uparrow$             & SSIM $\uparrow$  & LPIPS $\downarrow$ \\
MVSplat~\cite{chen2021mvsnerf}                          & 23.13         & 0.787 & \textbf{0.182} & ~          & 22.64          & 0.786 & 0.195 & 20.45           & 0.699 & 0.269 & 18.76            & 0.615  & 0.334 \\
\cite{chen2021mvsnerf} + ours    &  \textbf{23.17}           & \textbf{0.790} & \textbf{0.182} & \textbf{~} & \textbf{23.39}          & \textbf{0.812} & \textbf{0.180} & \textbf{22.29}           & \textbf{0.779} & \textbf{0.229} & \textbf{20.55}            & \textbf{0.707} & \textbf{0.319} \\
TranSplat~\cite{zhang2025transplat}                        & 23.22         & 0.787 & 0.180 & ~          & 22.75           & 0.787 & 0.192 & 20.66           & 0.706  & 0.263 & 18.48            & 0.598 & 0.348 \\
\cite{zhang2025transplat} + ours  &  \textbf{23.32}         &  \textbf{0.793} &  \textbf{0.179} & \textbf{~} & \textbf{23.65}          & \textbf{0.818} & \textbf{0.175} & \textbf{22.75}           & \textbf{0.792} & \textbf{0.216} & \textbf{21.30}            & \textbf{0.734} & \textbf{0.291} \\
DepthSplat~\cite{xu2025depthsplat}                      & \textbf{24.17}         & \textbf{0.819}  & \textbf{0.145} &            & 23.94           & 0.827 & 0.149  & 22.23            & 0.780  & 0.197  & 19.89            & 0.682 & 0.291 \\
\cite{xu2025depthsplat} + ours &  24.10         & 0.818  & 0.147 & \textbf{~} & \textbf{24.01}           & \textbf{0.832} & \textbf{0.147}  & \textbf{22.92}            & \textbf{0.811} & \textbf{0.181}  & \textbf{21.13}            & \textbf{0.756} & 0.\textbf{255 
}\end{tblr}
}
\label{tab:result_alpha_norm_noPretrained_dl3dv}
\end{table}

% \subsection{Optimization}
% \label{sub_sec:optimization}
% We train our model by utilizing both the standard color rendering loss and our proposed 3D sampling-based regularization. The final training loss $\mathcal{L}$ is as follow, 
% \begin{equation}
%     \mathcal{L} = \mathcal{L}_{\text{2D}}+\lambda\mathcal{L}_{\text{3D}}
% \end{equation}
% \begin{equation}
%     \mathcal{L}_{\text{2D}}=\text{MSE}(\mathbf{C}^{\text{2D}}, \mathbf{C}^{\text{gt}}) + \beta\, \text{LPIPS}(\mathbf{C}^{\text{2D}}, \mathbf{C}^{\text{gt }}),
% \end{equation}
% where $\mathbf{C}^{\text{2D}}$ can be obtained by stacking all the rendered color $\mathbf{c}(\mathbf{p})$ obtained in Eq.~\ref{}. It's worth noting that the rendering branch producing $\mathbf{C}^{\text{3D}}$ is only leveraged for a regularization during training. Af inference time, this branch can be fully discarded and our method still follows the standard rendering of 3DGS, thus maintaining its high rendering speed advantage. Additionally, the gradient from $\mathcal{L}_{\text{3D}}$ is only back-propagated to the predicted Gaussians scale $\mathbf{s}$ and the opacity $o^{\text{3D}}$, but not other Gaussians parameters. This helps to prevent the regularizer from altering other Gaussian parameters and to keep it as a pure scale regularizer.

\subsection{Optimization}
\label{sub_sec:optimization}
We train our model using both the standard color rendering loss and our proposed 3D sampling-based regularizer. The final training objective is,
\begin{equation}
    \mathcal{L} = (1-\lambda)\mathcal{L}_{\text{2D}}+\lambda\mathcal{L}_{\text{3D}}.
\end{equation}
\begin{equation}
    \mathcal{L}_{\text{2D}}=\text{MSE}(\mathbf{C}^{\text{2D}}, \mathbf{C}^{\text{gt}}) + \beta\, \text{LPIPS}(\mathbf{C}^{\text{2D}}, \mathbf{C}^{\text{gt}}),
\end{equation}
where $\lambda$ and $\beta$ are loss weights. It is worth noting that the rendering branch producing $\mathbf{C}^{\text{3D}}$ is used only as a regularizer during training. At inference time, this branch is discarded and our method follows the standard rendering pipeline of 3DGS~\cite{kerbl20233d}, thereby preserving its high rendering speed. Additionally, gradients from $\mathcal{L}_{\text{3D}}$ are back-propagated only to the predicted Gaussian scales $\mathbf{s}$ and the auxiliary opacity $o^{\text{3D}}$, but not to other Gaussian parameters. This helps prevent the regularizer from altering other Gaussian parameters and keeps it as a pure scale regularizer. Unless otherwise stated, alpha normalization is enabled by default during training.

% Explain generalizable 3DGS first. 

% Then explain each component that we introduce. 

% Alpha normalizatin
    % Observation

% 3D sampling-based regularization
    % 

% Optimization
    % 

% \input{main/tables/input_views_re10k}
% \input{main/tables/input_views_dl3dv}

\vspace{-4mm}
\section{Experiments}

\begin{figure}
\centering
\footnotesize

\begin{tblr}[]{
  width = \linewidth,
  colsep = 1pt,
  rowsep = 0.0pt,
  stretch = 0.25,
  row{2-Z} = {abovesep=1pt, belowsep=1pt},
  colspec = {Q[wd=4mm,c,m] X[c,m] X[c,m] X[c,m] X[c,m] X[c,m]},
  row{1} = {abovesep=2pt, belowsep=2pt},
  cell{1-Z}{1} = {halign=c, valign=m},
}

 & Ground-truth & 2 views & 4 views & 8 views & 16 views \\

% ===================== Sample 1 =====================

\raisebox{10pt}{\rotlabel{MVSplat~\cite{chen2024mvsplat}}} &
\figimg{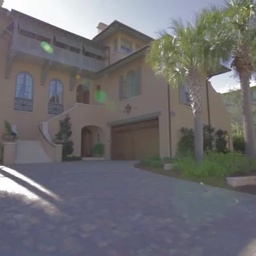} &
\figimg{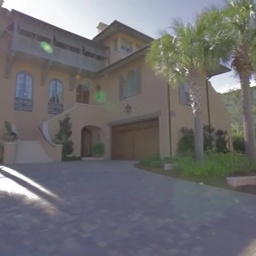} &
\figimg{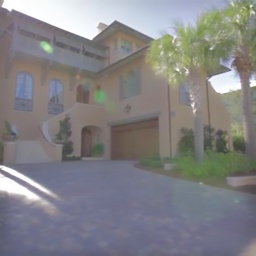} &
\figimg{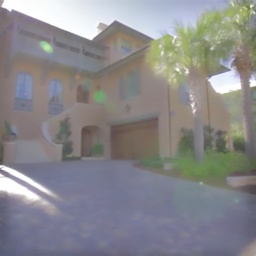} & 
\figimg{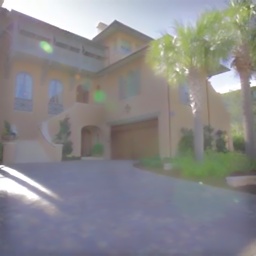} \\

\raisebox{10pt}{\rotlabel{$\alpha$-MVSplat}} &
\figimg{main/Figures/alpha_norm_vis/de129b4aa11af575/gt.jpg} &
\figimg{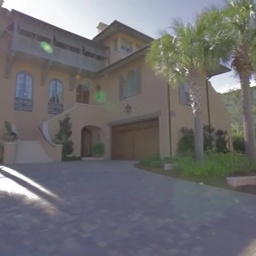} &
\figimg{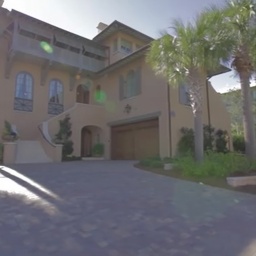} &
\figimg{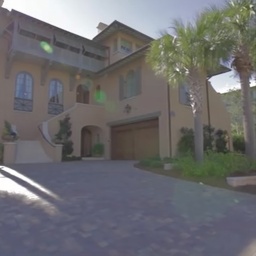} & 
\figimg{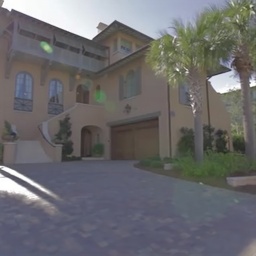} \\

\hline[dashed, wd=0.5pt]

% ===================== Sample 2 =====================

\raisebox{30pt}{\rotlabel{DepthSplat~\cite{xu2025depthsplat}}} &
\figimg{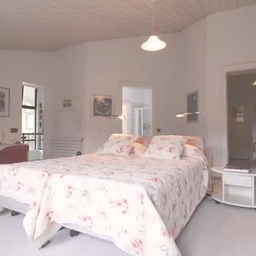} &
\figimg{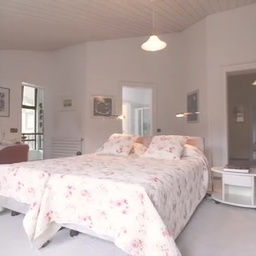} &
\figimg{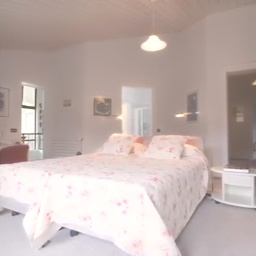} &
\figimg{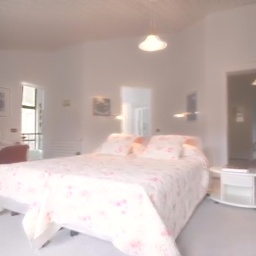} & 
\figimg{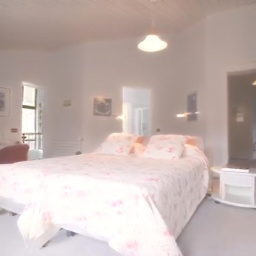} \\

\raisebox{10pt}{\rotlabel{$\alpha$-DepthSplat}} &
\figimg{main/Figures/alpha_norm_vis/1961bb85524de229/gt.jpg} &
\figimg{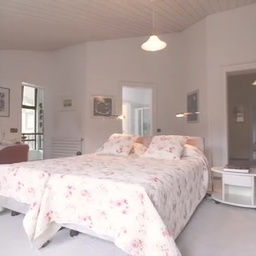} &
\figimg{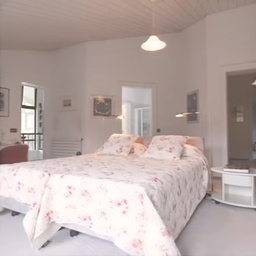} &
\figimg{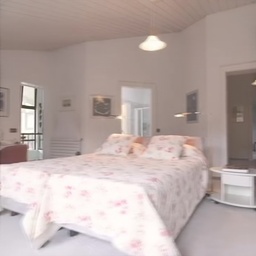} & 
\figimg{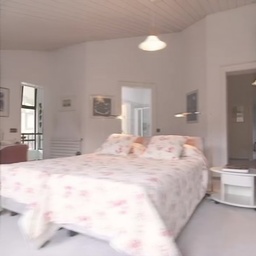}

\end{tblr}
\caption{\textbf{Qualitative results on RealEstate10K}. Integrating alpha normalization helps to mitigate the overbrightness issues encountered by \cite{chen2024mvsplat, xu2025depthsplat} when increasing the number of input views.}
\label{fig:qualitative_result_alpha_norm}
\end{figure}
\begin{table}[t]
\centering
\scriptsize
\caption{\textbf{Comparision on DL3DV under high-resolution rendering}. All models are trained with $256\times448$ images and render outputs at up to $8\times$ the training resolution at test time. Our method significantly outperform DepthSplat on high-resolution renderings.}
\resizebox{\textwidth}{!}{
\begin{tblr}{
  colsep  = 1.5pt,
  rowsep  = 0.5pt,
  column{even} = {r},
  column{3} = {r},
  column{7} = {r},
  column{9} = {r},
  column{11} = {r},
  column{13} = {r},
  cell{1}{2} = {c=3}{c},
  cell{1}{5} = {c},
  cell{1}{6} = {c=3}{c},
  cell{1}{9} = {c=3}{c},
  cell{1}{12} = {c=3}{c},
  cell{2}{2} = {c},
  cell{2}{3} = {c},
  cell{2}{4} = {c},
  cell{2}{5} = {c},
  cell{2}{6} = {c},
  cell{2}{7} = {c},
  cell{2}{8} = {c},
  cell{2}{9} = {c},
  cell{2}{10} = {c},
  cell{2}{11} = {c},
  cell{2}{12} = {c},
  cell{2}{13} = {c},
  cell{2}{14} = {c},
  % cell{3}{2} = {font=\bfseries},
  % cell{3}{3} = {font=\bfseries},
  % cell{3}{4} = {font=\bfseries},
  cell{3}{5} = {c},
  % cell{4}{2} = {font=\bfseries},
  % cell{4}{3} = {font=\bfseries},
  % cell{4}{4} = {font=\bfseries},
  cell{4}{5} = {c},
  % cell{4}{6} = {font=\bfseries},
  % cell{4}{7} = {font=\bfseries},
  % cell{4}{8} = {font=\bfseries},
  % cell{4}{9} = {font=\bfseries},
  % cell{4}{10} = {font=\bfseries},
  % cell{4}{11} = {font=\bfseries},
  % cell{4}{12} = {font=\bfseries},
  % cell{4}{13} = {font=\bfseries},
  % cell{4}{14} = {font=\bfseries},
  % cell{5}{2} = {font=\bfseries},
  % cell{5}{3} = {font=\bfseries},
  % cell{5}{4} = {font=\bfseries},
  cell{5}{5} = {c},
  % cell{6}{2} = {font=\bfseries},
  % cell{6}{3} = {font=\bfseries},
  % cell{6}{4} = {font=\bfseries},
  % cell{6}{6} = {font=\bfseries},
  % cell{6}{7} = {font=\bfseries},
  % cell{6}{8} = {font=\bfseries},
  % cell{6}{9} = {font=\bfseries},
  % cell{6}{10} = {font=\bfseries},
  % cell{6}{11} = {font=\bfseries},
  % cell{6}{12} = {font=\bfseries},
  % cell{6}{13} = {font=\bfseries},
  % cell{6}{14} = {font=\bfseries},
  % cell{7}{2} = {font=\bfseries},
  % cell{7}{3} = {font=\bfseries},
  % cell{7}{4} = {font=\bfseries},
  % cell{8}{2} = {font=\bfseries},
  % cell{8}{3} = {font=\bfseries},
  % cell{8}{4} = {font=\bfseries},
  % cell{8}{6} = {font=\bfseries},
  % cell{8}{7} = {font=\bfseries},
  % cell{8}{8} = {font=\bfseries},
  % cell{8}{9} = {font=\bfseries},
  % cell{8}{10} = {font=\bfseries},
  % cell{8}{11} = {font=\bfseries},
  % cell{8}{12} = {font=\bfseries},
  % cell{8}{13} = {font=\bfseries},
  % cell{8}{14} = {font=\bfseries},
  vline{2,6} = {1}{Black},
  vline{3,7,10,13} = {1}{},
  vline{2,5-6,9,12} = {1-9}{Black},
  hline{3,5,7} = {1,4,6,8,11,14}{},
  hline{3,5,7} = {2-3,7,9-10,12-13}{Black},
}
           & $256\times 448$ &        &       &   & $512\times 896$  &       &        & $1024\times 1792$  &       &        & $2048\times 3584$  &       &        \\
           & PSNR $\uparrow$        & SSIM $\uparrow$  & LPIPS $\downarrow$ & ~ & PSNR $\uparrow$           & SSIM $\uparrow$ & LPIPS $\downarrow$ & PSNR $\uparrow$            & SSIM $\uparrow$ & LPIPS $\downarrow$ & PSNR $\uparrow$           & SSIM $\uparrow$  & LPIPS  $\downarrow$ \\
MVSplat~\cite{chen2021mvsnerf}                          & 23.13         & 0.787 & 0.182 & ~          & 21.15          & 0.713 & 0.302 & 19.24           & 0.649 & 0.429 & 18.34            & 0.661  & 0.494 \\
\cite{chen2021mvsnerf} + ours    &  \textbf{23.17}         &  \textbf{0.790} &  \textbf{0.181} & \textbf{~} & \textbf{21.42}          & \textbf{0.724} & \textbf{0.299} & \textbf{20.16}           & \textbf{0.681} & \textbf{0.414} & \textbf{19.50}            & \textbf{0.698} & \textbf{0.481} \\
TranSplat~\cite{zhang2025transplat}                        & 23.22         & 0.788 & 0.180 & ~          & 21.22          & 0.715 & 0.299 & 19.30           & 0.651  & 0.425 & 17.73            & 0.665 & 0.486 \\
\cite{zhang2025transplat} + ours  &  \textbf{23.32}         &  \textbf{0.793} &  \textbf{0.179} & \textbf{~} & \textbf{21.55}          & \textbf{0.726} & \textbf{0.298} & \textbf{20.16}           & \textbf{0.685} & \textbf{0.407} & \textbf{19.34}            & \textbf{0.690} & \textbf{0.486} \\
DepthSplat~\cite{xu2025depthsplat}                      & \textbf{24.17} & \textbf{0.819} & \textbf{0.145} & & 20.82 & 0.722 & 0.284 & 18.18 & 0.626 & 0.438 & 17.02 & 0.626 & 0.505 \\
\cite{xu2025depthsplat} + ours &  24.10 & 0.818 & 0.147 & & \textbf{21.79} & \textbf{0.742} & \textbf{0.265} & \textbf{20.18} & \textbf{0.687} & \textbf{0.392} & \textbf{19.38} & \textbf{0.696} & \textbf{0.471} 
\end{tblr}
}
\label{tab:high_resolution_comparision}
\end{table}

\subsection{Experimental Setup}
\noindent\textbf{Dataset.} We validate our method on RealEstate10K~\cite{zhou2018stereo} and DL3DV~\cite{ling2024dl3dv} for the NVS task, following the train/test splits used in prior work~\cite{chen2021mvsnerf, xu2025depthsplat}. On RealEstate10K, all models are trained with two input images at a resolution of $256\times256$, and evaluated with up to 16 input views during inference. On the DL3DV dataset, following~\cite{xu2025depthsplat}, we train the model with 2--6 input views at $256\times448$ resolution and evaluate under two settings: (i) varying number of input views at inference time (up to 48) and (ii) high-resolution rendering (up to $8\times$ the training resolution). 

\noindent\textbf{Baselines.} We use MVSplat~\cite{chen2024mvsplat}, Transplat~\cite{zhang2025transplat} and DepthSplat~\cite{xu2025depthsplat} as baseline methods in our experiments and validate the effectiveness of our proposed modules through the performance gains obtained when integrating them into the baselines. 
% For evaluation, we adopt the standard NVS evaluation metrics including PSNR, SSIM~\cite{wang2004image} and LPIPS~\cite{zhang2018unreasonable} to measure the quality of rendered images.

\vspace{1mm}
% Implementation details 
\noindent\textbf{Implementation Details. }
% We adopt DepthSplat~\cite{xu2025depthsplat} as a backbone architecture and integrate our proposed alpha normalization and 3D sampling-based regularizer on top of this method. 
Following \cite{xu2025depthsplat}, we train our method for $150,000$ iterations using a batch size of 32, the number of training iterations and batch size on the DL3DV dataset are $100,000$ and $8$, respectively. We use Adam optimizer with a learning rate schedule similar to \cite{xu2025depthsplat}. In addition, we set the reference number of overlapping Gaussians $m=1$ (see Eq.~\ref{eq:alphaNormalisation}), depth error threshold $\tau=0.5$ and the loss weights $\lambda=\beta=0.05$. All training is conducted using 8 RTX 4090 GPUs.

% \noindent\textbf{. } We use MVSplat~\cite{chen2024mvsplat}, Transplat~\cite{zhang2025transplat} and DepthSplat~\cite{xu2025depthsplat} as baseline methods and validate the effectiveness of our proposed modules through the performance gains obtained when integrating them into the baselines. 
% Additionally, to demonstrate that alpha normalization is architecture-agnostic, we apply it to pretrained models of several the baseline methods, including MVSplat~\cite{chen2024mvsplat}, TranSplat~\cite{zhang2025transplat}, and DepthSplat~\cite{xu2025depthsplat}, and compare the performance before and after adding this module.

% Build upon DepthSplat~\cite{xu2025depthsplat}
% Training time, iteration, GPU: 8x RTX4090. Learnng rate
% We set the reference number of overlapping Gaussians as $m=2$, Loss weight, depth error threshold. 
% we also train our method on for the 

\vspace{-3mm}
\subsection{Results}

% \subsection{Generalization to Varying Number of Input Views} 
\noindent\textbf{Varying Numbers of Input Views.} In Tab.~\ref{tab:result_alpha_norm_noPretrained_re10k}, we apply alpha normalization to pretrained models of existing methods~\cite{chen2024mvsplat, zhang2025transplat, xu2025depthsplat} \textbf{without retraining} and compare them with their original models. All methods are pretrained using 2 input views on the RealEstate10k dataset and evaluated with up to 16 input views. 
% In addition, we activate alpha normalization only when the number of input views is greater than 2, since the models are already trained to produce plausible outputs for the 2-view setting. 
The results demonstrates that, without normalization, the performance of prior works drops significantly as the number of input views increases. In contrast, integrating alpha normalization on average improves the PSNR by \textbf{3.3} dB, \textbf{5.53} dB, and \textbf{5.71} dB in the 4-view, 8-view, and 16-view settings, respectively. Tab.~\ref{tab:result_alpha_norm_noPretrained_re10k} also shows that our alpha normalization achieves large improvements and outperforms GNN~\cite{zhang2024gaussian} across all settings. Unlike GNN~\cite{zhang2024gaussian}, our method preserves the total number of Gaussians without fusing them, and is thus less sensitive to finding corresponding Gaussians across views. We show the qualitative comparison in Fig.~\ref{fig:qualitative_result_alpha_norm}. It can be observed that with an increase in the number of input views, existing works~\cite{chen2024mvsplat, xu2025depthsplat} tend to render brighter images. Alpha normalization helps to alleviate this over-brightness issue, and thereby preserving the image quality. 

Additionally, we introduce our proposed alpha normalization and 3D sampling-based regularizer into the training process of the baselines and train them on the DL3DV dataset. During training, all methods take 2 to 6 inputs images. Tab.~\ref{tab:result_alpha_norm_noPretrained_dl3dv} shows our method achieves similar performance to baselines under the same configuration during training (i.e., 6 input views), while outperforming them when the number of input views exceeds 6. Moreover, the performance gain of our method increases with the number of input views, indicating stronger robustness to varying input-view counts.

\begin{figure}[t]
    \centering
    % Set to 0pt if you want absolutely no gap between the images
    \setlength{\tabcolsep}{1pt} 
    
    \begin{tabular}{ccc}
        % --- Header Row ---
        Ground-truth & DepthSplat~\cite{xu2025depthsplat} & Ours \\

        % --- Row 1 Images ---
        \includegraphics[width=0.32\linewidth]{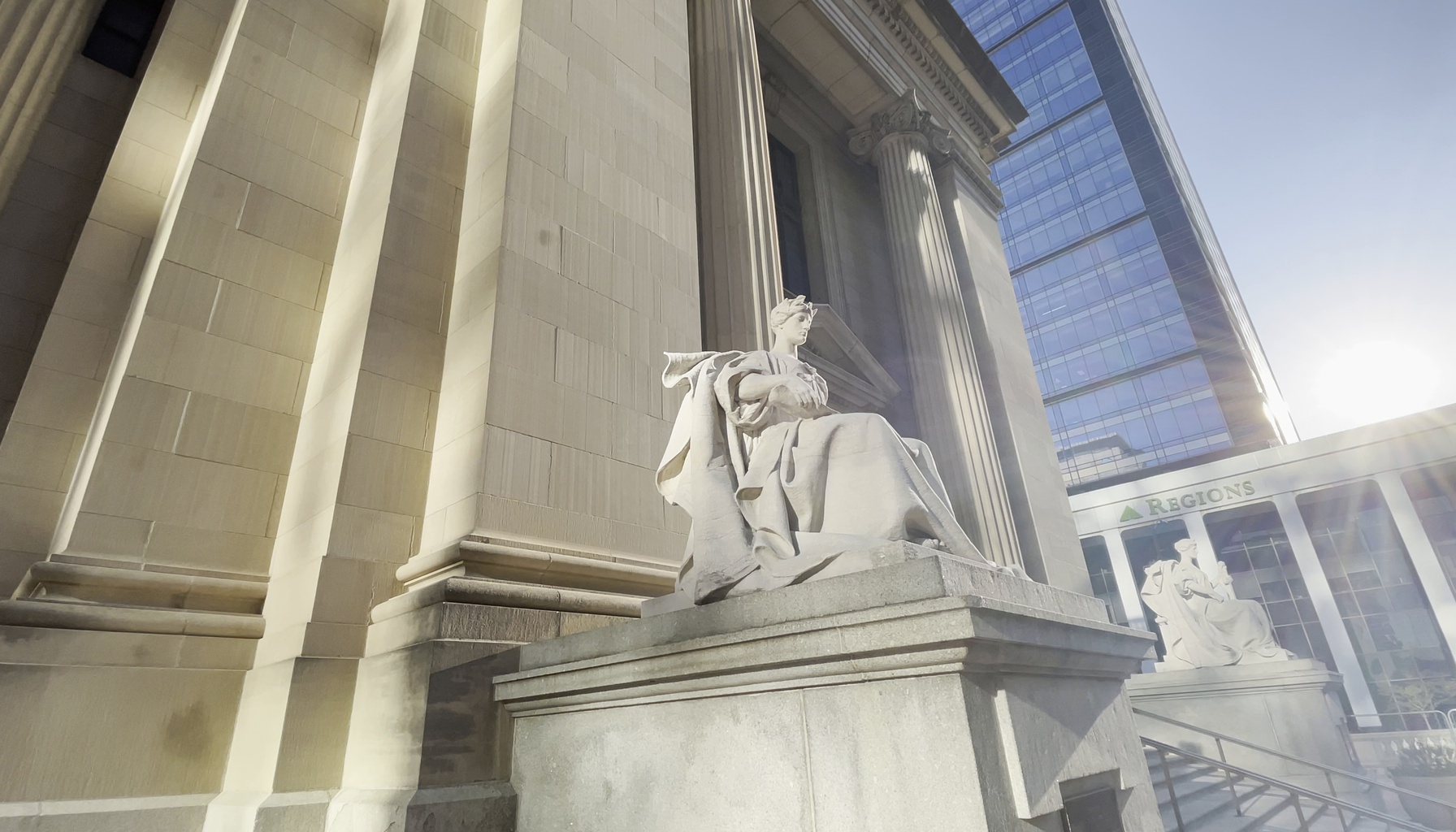} & 
        \includegraphics[width=0.32\linewidth]{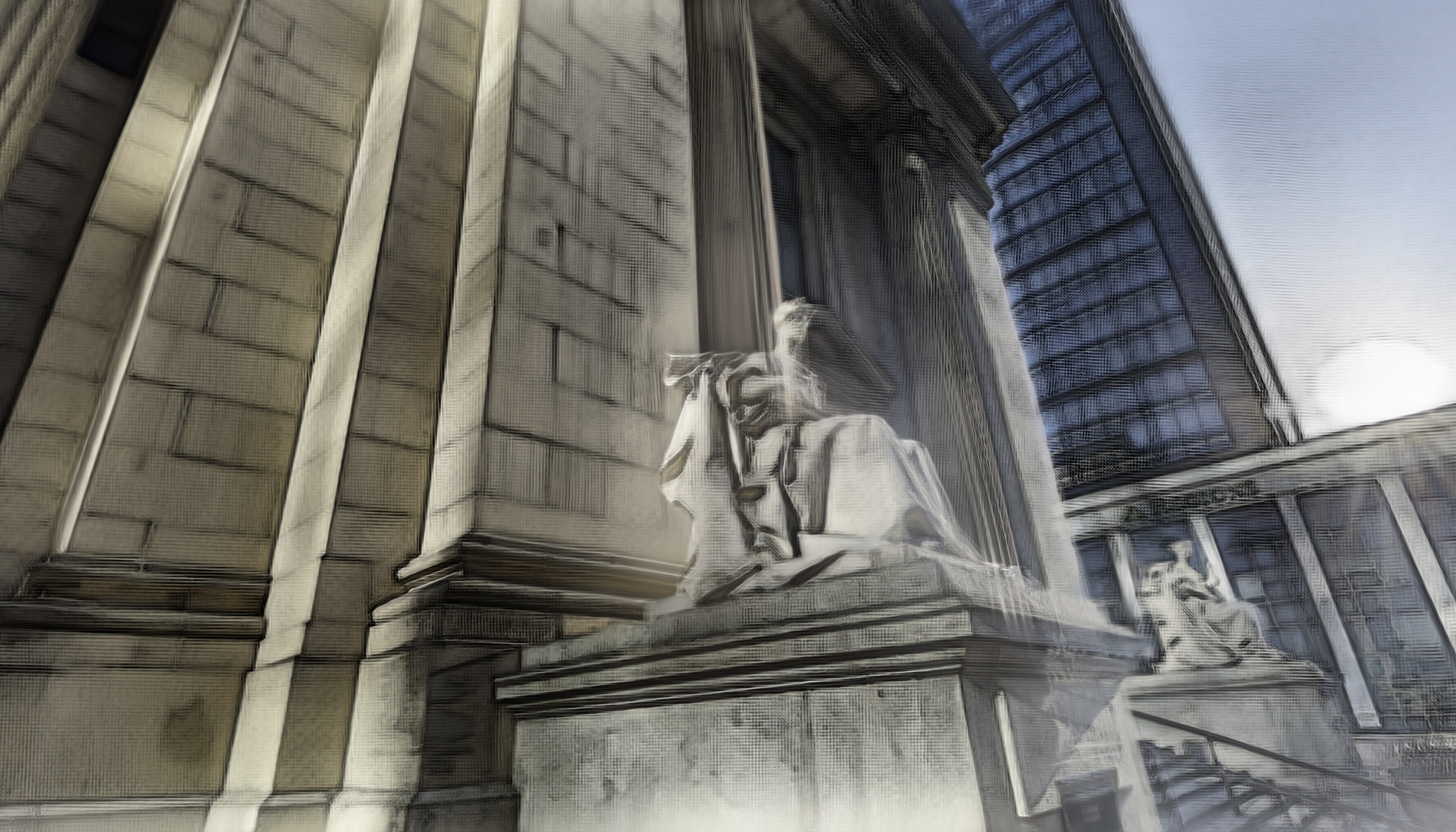} & 
        \includegraphics[width=0.32\linewidth]{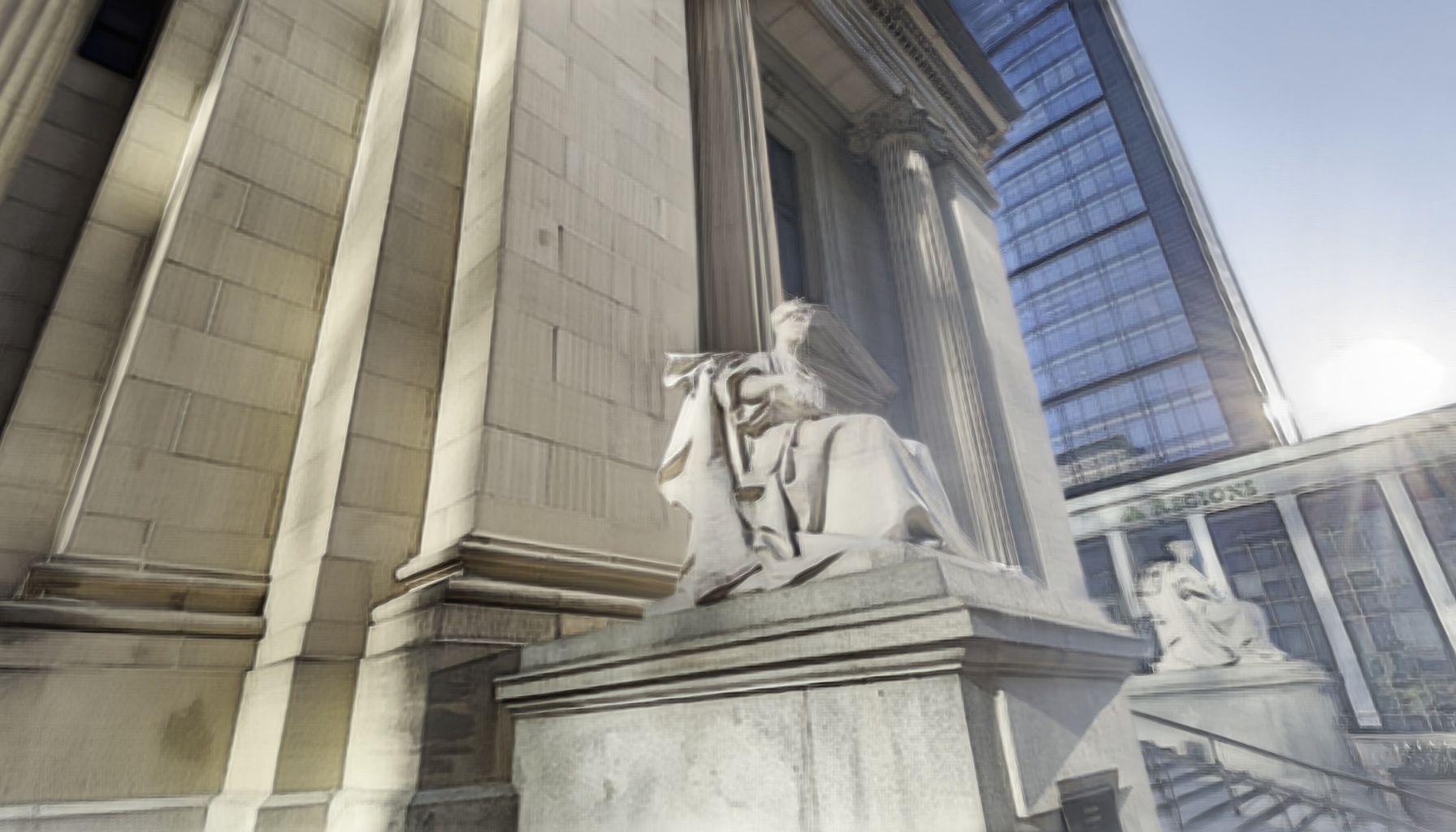} \\

        \includegraphics[width=0.32\linewidth]{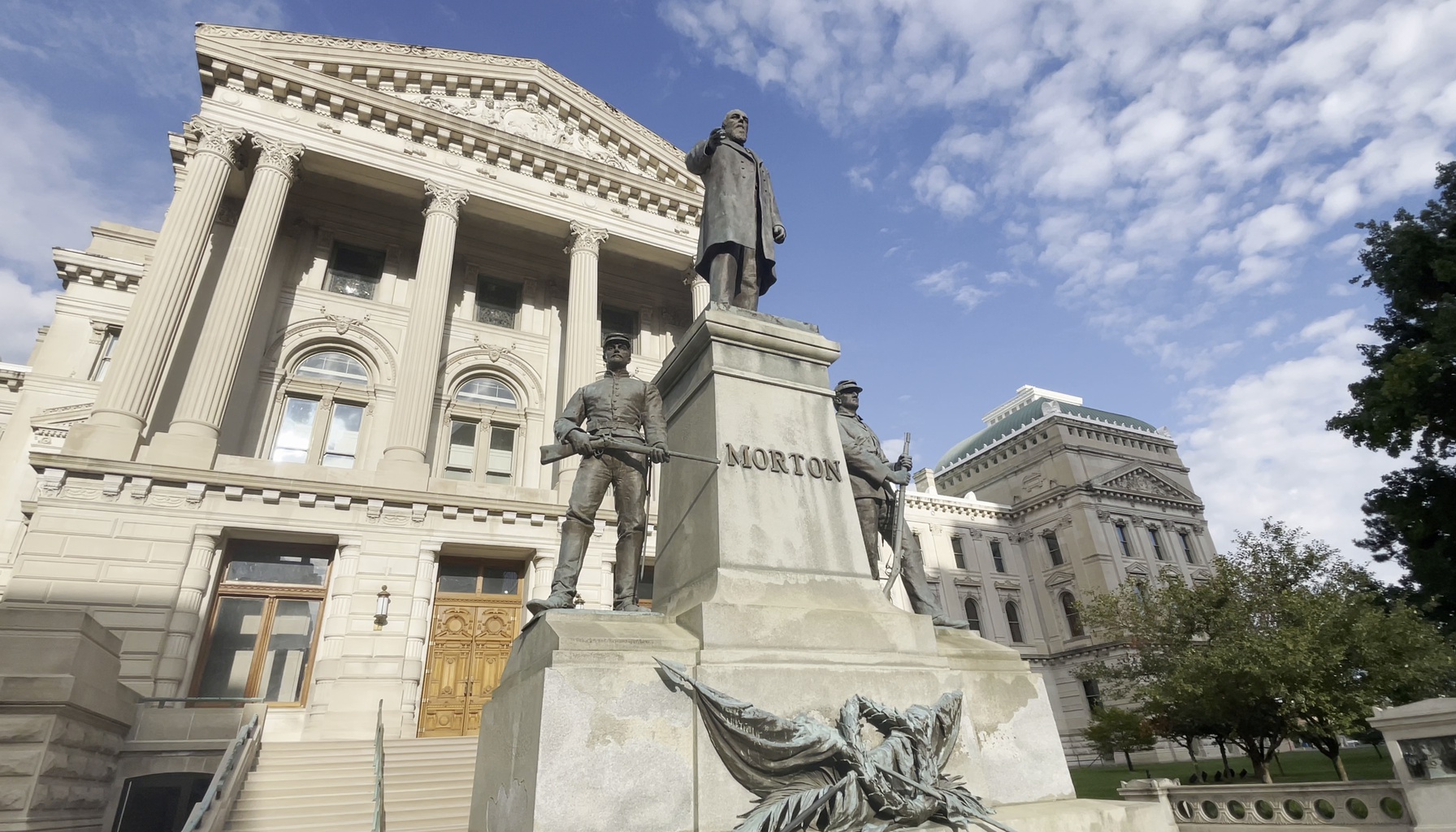} & 
        \includegraphics[width=0.32\linewidth]{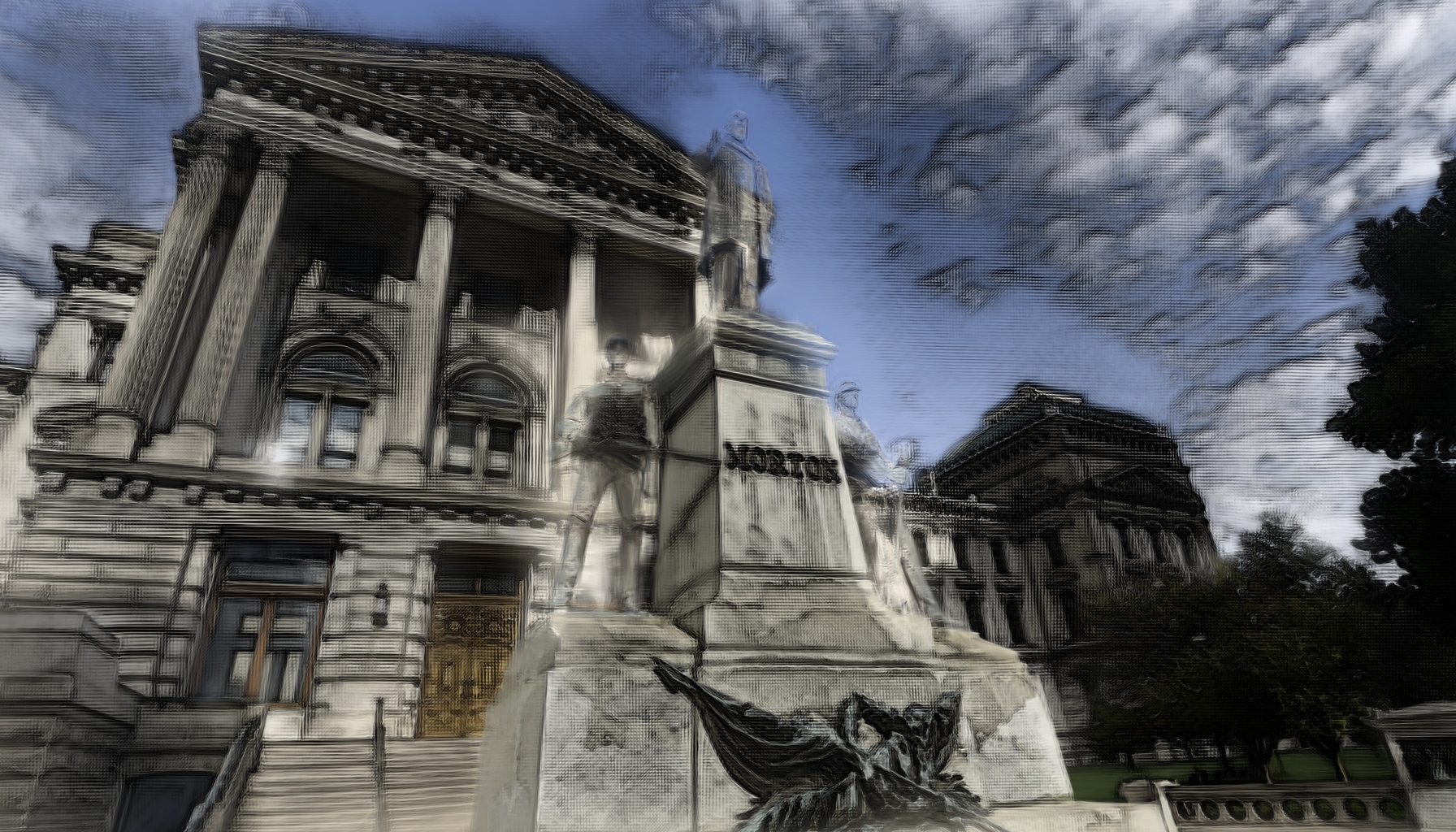} & 
        \includegraphics[width=0.32\linewidth]{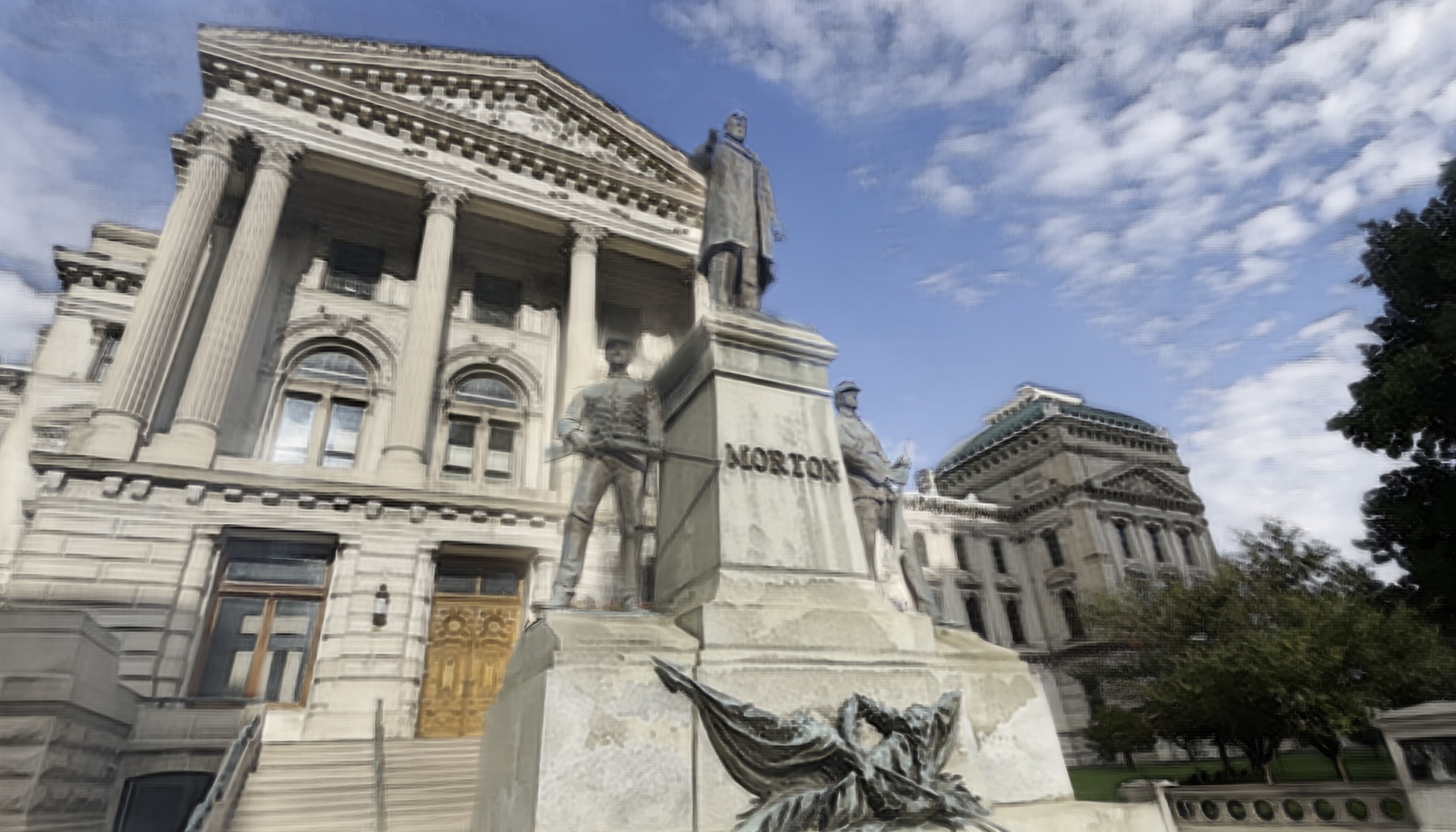} \\

        % --- Row 2 Images ---
        \includegraphics[width=0.32\linewidth]{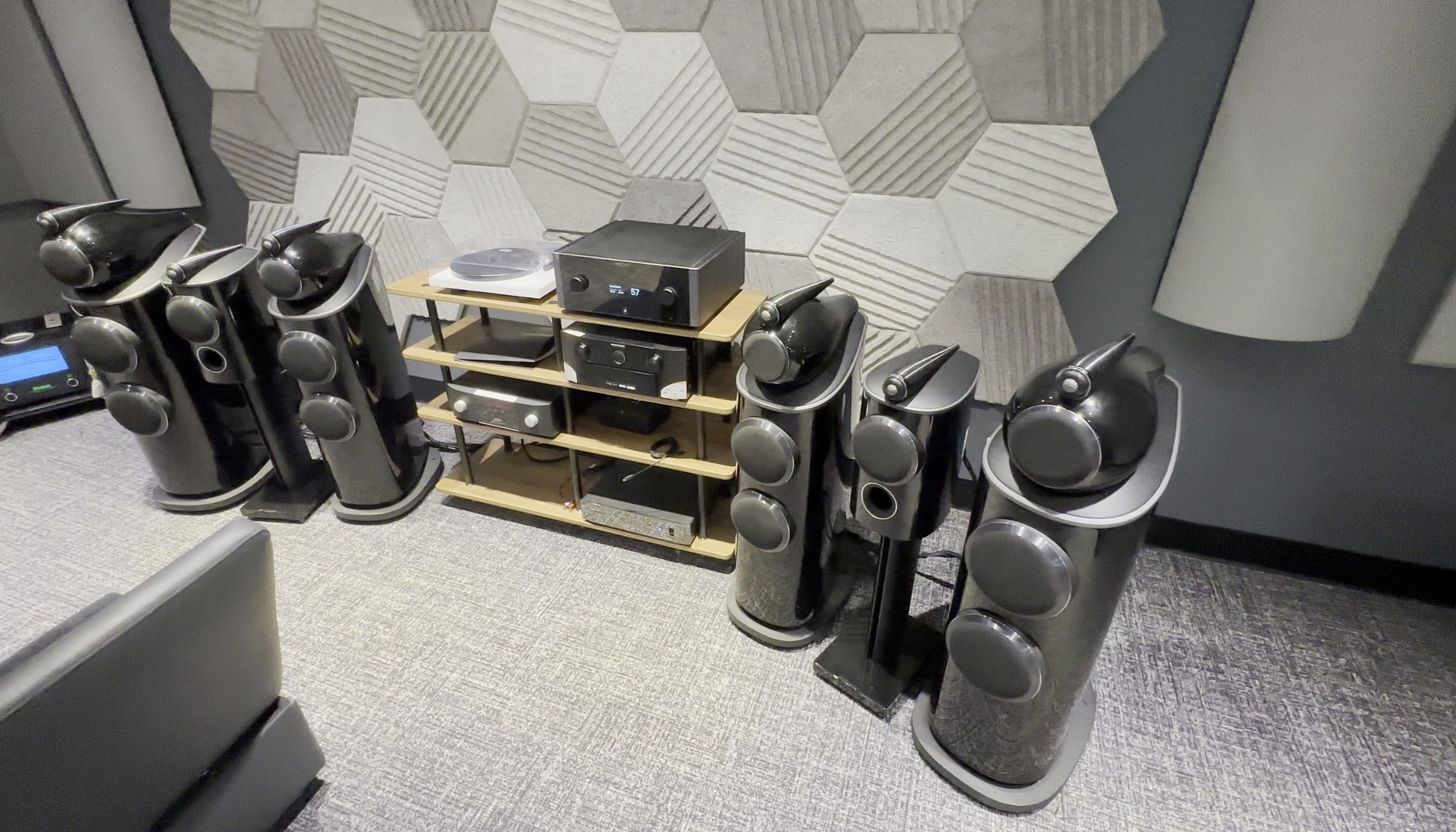} & 
        \includegraphics[width=0.32\linewidth]{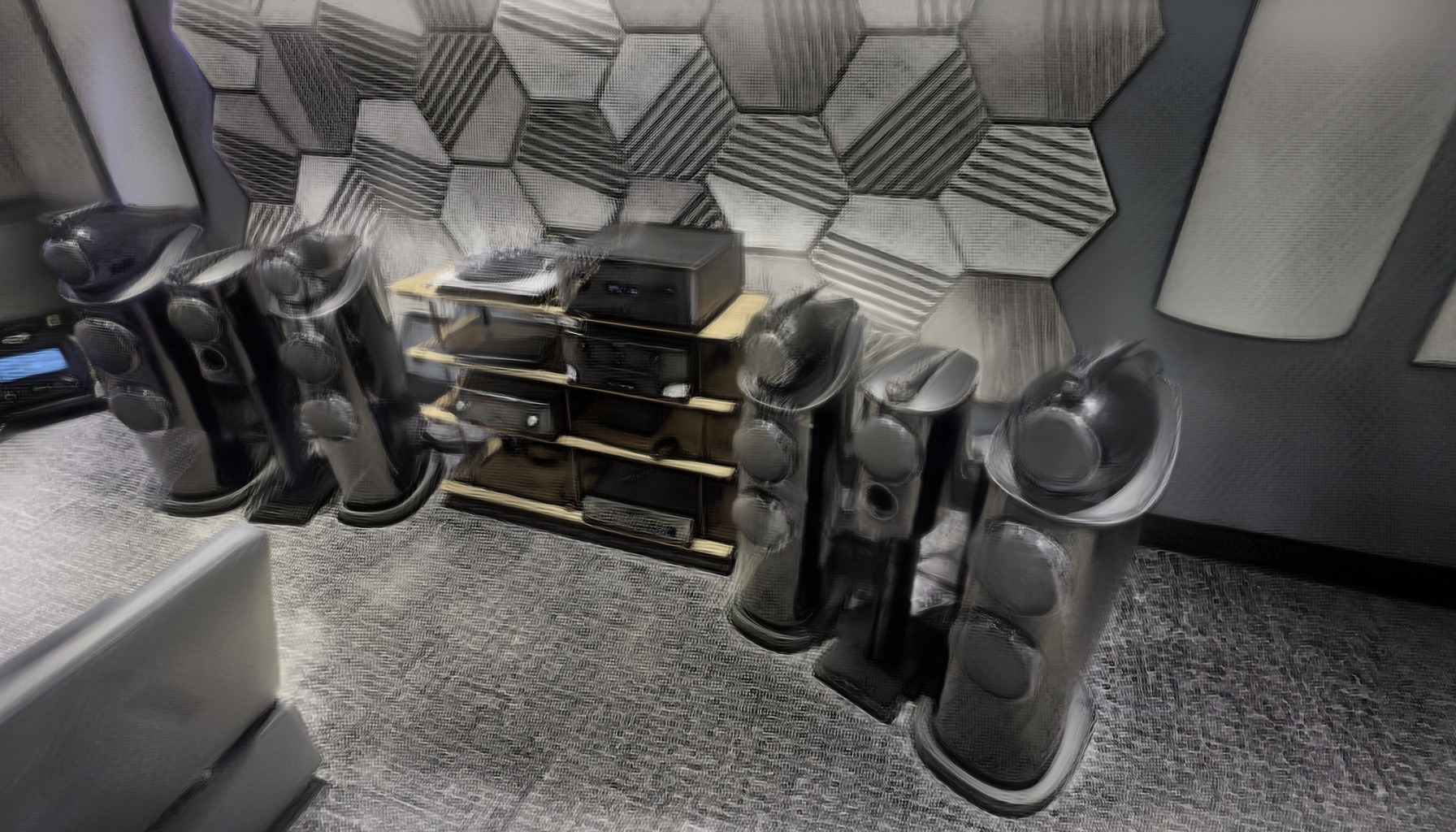} & 
        \includegraphics[width=0.32\linewidth]{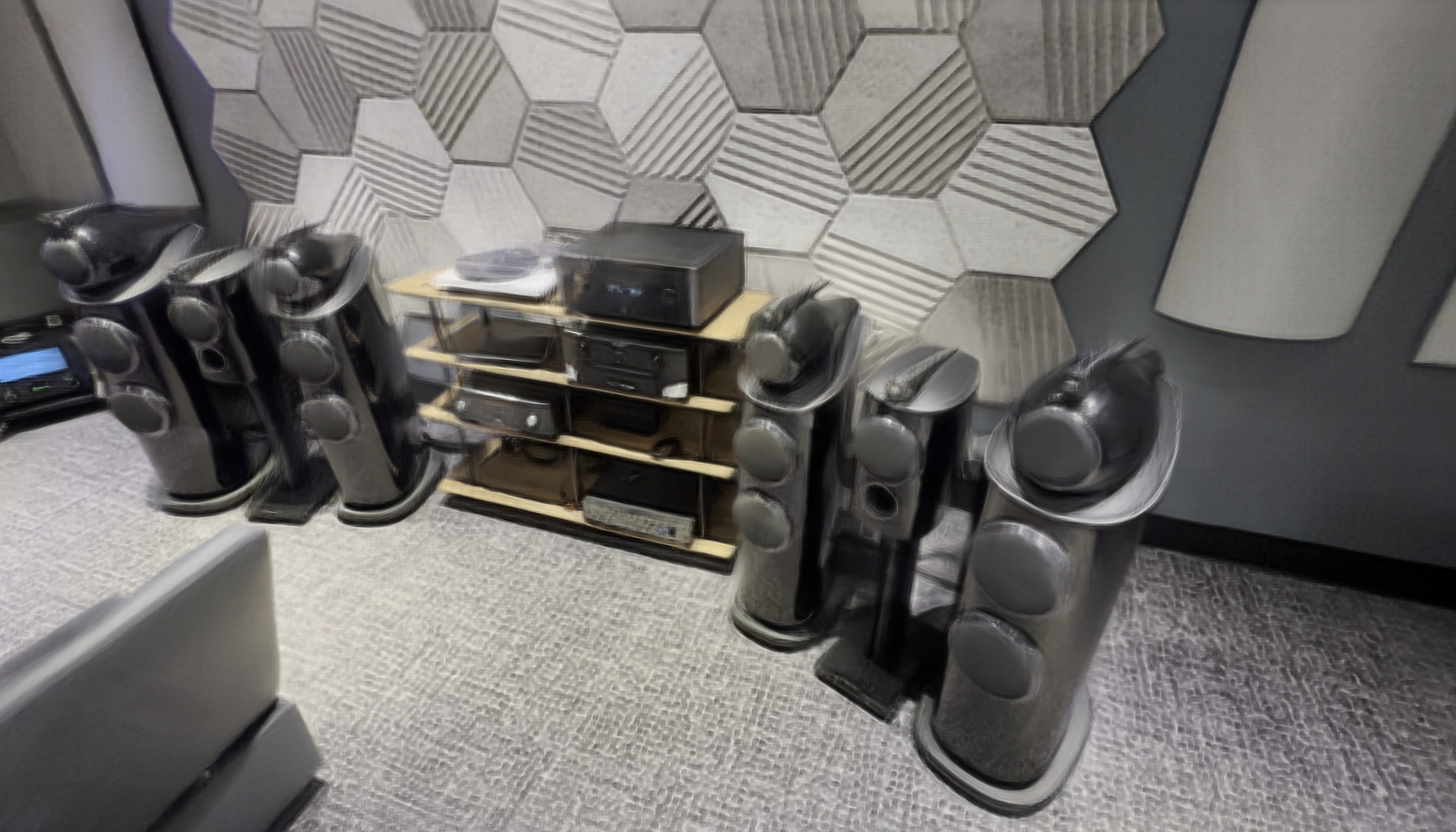} \\

    \end{tabular}
    
    \caption{\textbf{High-resolution rendering on DL3DV dataset}. Our method significantly alleviates the holes issues exhibited by DepthSplat~\cite{xu2025depthsplat}. }
    \label{fig:qualitative_result_high_resolution}
\end{figure}

\begin{table}[t]
\centering
\footnotesize
\caption{\textbf{Ablation study on DL3DV}. Integrating alpha normalization, 3D-sampling based regularizer improve the image quality under varying input view and high-resolution rendering, respectively.  }
% \resizebox{\textwidth}{!}{
\begin{tblr}{
  colsep  = 1.5pt,
  rowsep  = 1pt,
  columns = {c},
  % Define row/column spans
  cell{1}{1} = {r=2}{c,m}, % Alpha norm. (spans 2 rows, centered middle)
  cell{1}{2} = {r=2}{c,m}, % 3D-based regu. (spans 2 rows, centered middle)
  cell{1}{3} = {c=3}{c},   % Group 1 header (spans 3 cols)
  cell{1}{6} = {c=3}{c},   % Group 2 header (spans 3 cols)
  cell{1}{9} = {c=3}{c},   % Group 3 header (spans 3 cols)
  % Vertical lines
  vline{1,2,3,6,9,12}  = {solid},  % Solid lines separating main groups
  % Horizontal lines
  hline{3}     = {1-11}{solid},      % Solid line below sub-headers
}
{Alpha\\norm.} & 
{3D-sampling\\based regu.} & 
6 views - $256 \times 448$ & & & 
24 views - $256 \times 448$ & & & 
6 views - $2048 \times 3584$ & & \\

& & 
PSNR $\uparrow$ & SSIM $\uparrow$ & LPIPS $\downarrow$ & 
PSNR $\uparrow$ & SSIM $\uparrow$ & LPIPS $\downarrow$ & 
PSNR $\uparrow$ & SSIM $\uparrow$ & LPIPS $\downarrow$ \\

& & 
\underline{24.17} & \underline{0.819} & \textbf{0.145} & 
22.23 & 0.780 & 0.197 & 
17.02 & 0.626 & 0.505 \\

\checkmark & & 
\textbf{24.21} & \textbf{0.820} & \underline{0.146} & 
\textbf{23.18} & \textbf{0.813} & \textbf{0.181} & 
\underline{18.11} & \underline{0.661} & \underline{0.490} \\

\checkmark & \checkmark & 
24.10 & 0.818 & 0.147 & 
\underline{22.92} & \underline{0.811} & \textbf{0.181} & 
\textbf{19.38} & \textbf{0.696} & \textbf{0.471} \\
\end{tblr}
% }
\label{tab:ablation_study}
\end{table}

\noindent\textbf{High-Resolution Rendering. } Tab.~\ref{tab:high_resolution_comparision} compares our methods with existing works on high-resolution rendering. All methods are trained at a resolution of $256 \times 448$ and evaluated by rendering images at up to $8\times$ higher resolution. Integrating our proposed methods into the baselines consistently improve the image quality, with the performance gain increase proportionally to the image resolution. For the PSNR metric, on average, our methods outperforms the baseline models by \textbf{0.52} dB, \textbf{1.26} dB and \textbf{1.71} dB while rendering images at $2\times$, $4\times$ and $8\times$ higher resolution, respectively. The improved performance benefits from more accurate Gaussian scale estimation: existing works tends to predict overly small-scale Gaussians, causing holes in the rendered images. In contrast, our 3D sampling-based regularizer encourages Gaussians with sufficiently large scales to cover surface regions visible in input views, thereby alleviating the holes artifacts. 
This is further supported by the qualitative comparison in Fig.~\ref{fig:qualitative_result_high_resolution}, showing that DepthSplat exhibits noticeable holes in its images while our method yields more complete renderings. 

\noindent\textbf{Additional Results.} We provide more qualitative results, cross-dataset evaluation, zoom-in rendering evaluation and more analysis in the supplementary material.

% \textcolor{red}{Mention cross-dataset evaluation in supps.}

% \input{main/4_experiments_subsections/cross_dataset_generalization}

\vspace{-3mm}
\subsection{Ablation Study}

% In this section, we conduct ablation studies to illustrate the impact of each proposed module. The results are shown in Tab.~\ref{tab:ablation_study}. 

\begin{figure}
\centering
\footnotesize

\begin{tblr}[
]{
  colsep = 1pt,
  rowsep = 0.0pt, % Set explicitly to 0pt
  stretch = 0.25,  % Turns off default line stretching
  row{2-Z} = {abovesep=0pt, belowsep=0pt}, % Kills inner cell padding for all image rows
  colspec = {Q[wd=4mm,c,m] X[c,m] X[c,m] X[c,m] X[c,m]},
  row{1} = {abovesep=2pt, belowsep=2pt}, % Keeps a normal gap for your text header
  cell{1-Z}{1} = {halign=c, valign=m}, 
  % hline{5} = {dashed, abovesep=0pt, belowsep=0pt},
}

 & Ground-truth & DepthSplat~\cite{xu2025depthsplat} & With $ \alpha $ norm. & Full model \\

% \smash{\rotatebox[origin=c]{90}{2 views}} &  
% \includegraphics[width=\linewidth]{main/Figures/alpha_norm_vis/mvsplat_rendered_2views_TrueAlpha.png} &
% \includegraphics[width=\linewidth]{main/Figures/alpha_norm_vis/depthsplat_rendered_2views_TrueAlpha.png} &
% \includegraphics[width=\linewidth]{main/Figures/alpha_norm_vis/transplat_rendered_2views_TrueAlpha.png} &
% \includegraphics[width=\linewidth]{main/Figures/alpha_norm_vis/gt.png} \\

\raisebox{22.5pt}{\smash{\rotatebox[origin=c]{90}{More views}}} &  
\includegraphics[width=\linewidth]{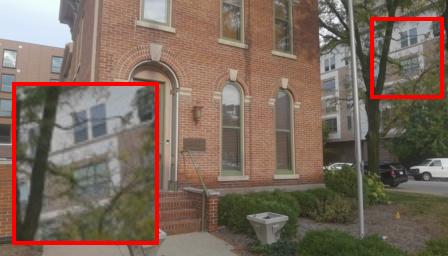} &
\includegraphics[width=\linewidth]{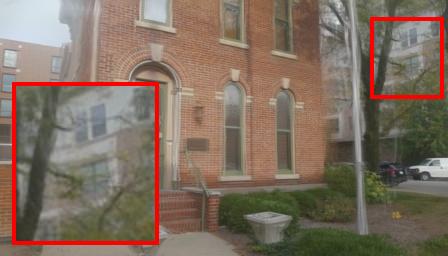} &
\includegraphics[width=\linewidth]{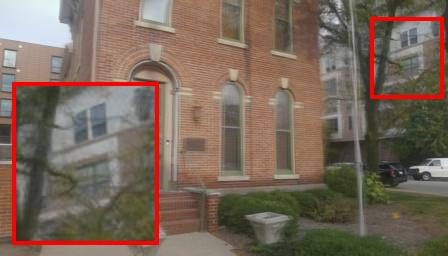} &
\includegraphics[width=\linewidth]{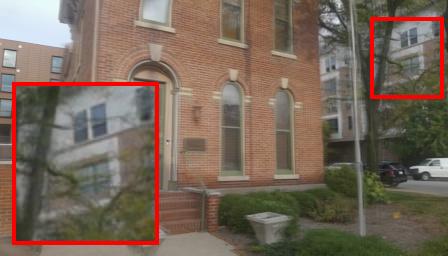} \\

\SetRow{belowsep=2pt}
\raisebox{20pt}{\smash{\rotatebox[origin=c]{90}{High reso.}}} &  
\includegraphics[width=\linewidth]{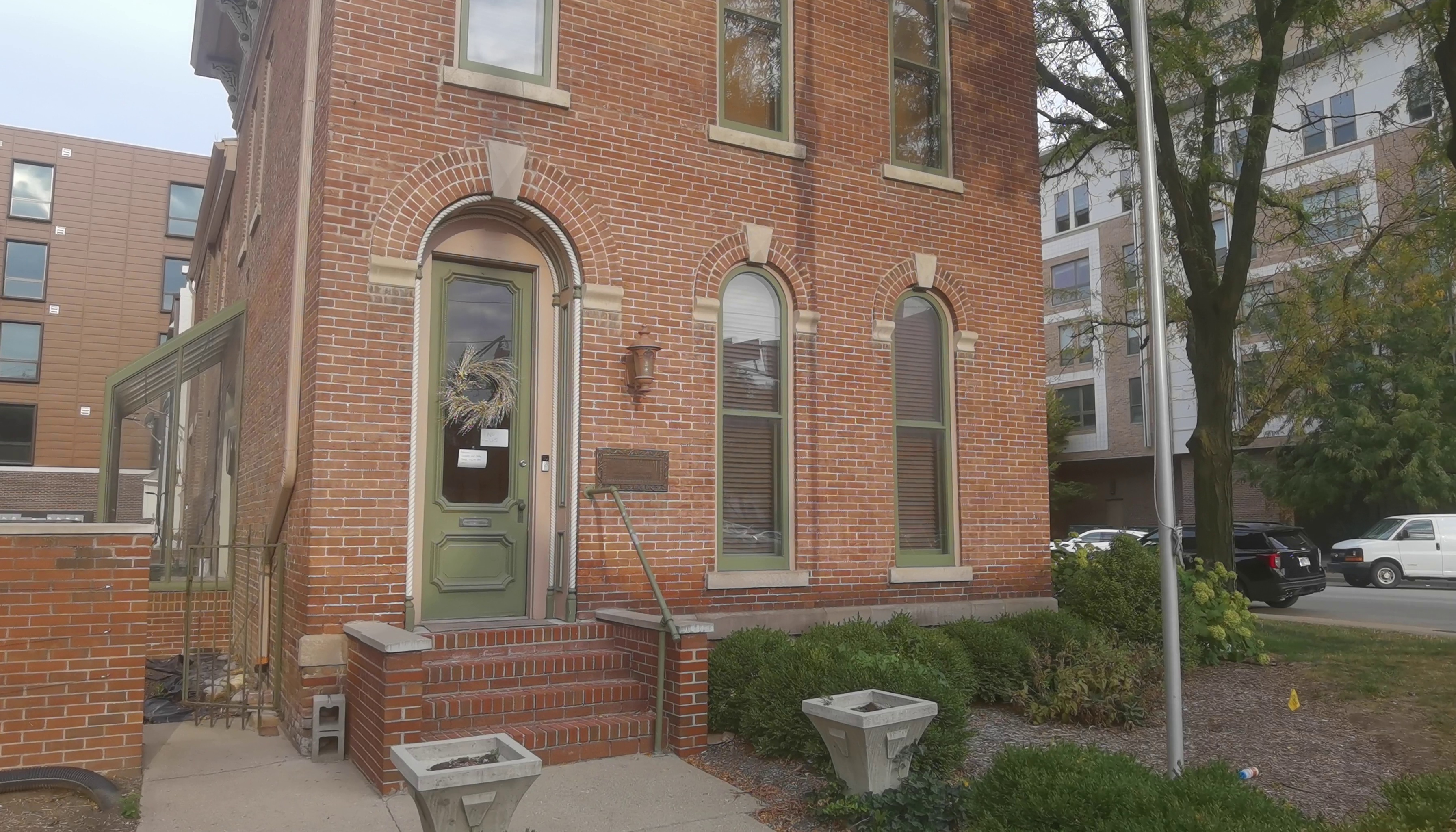} &
\includegraphics[width=\linewidth]{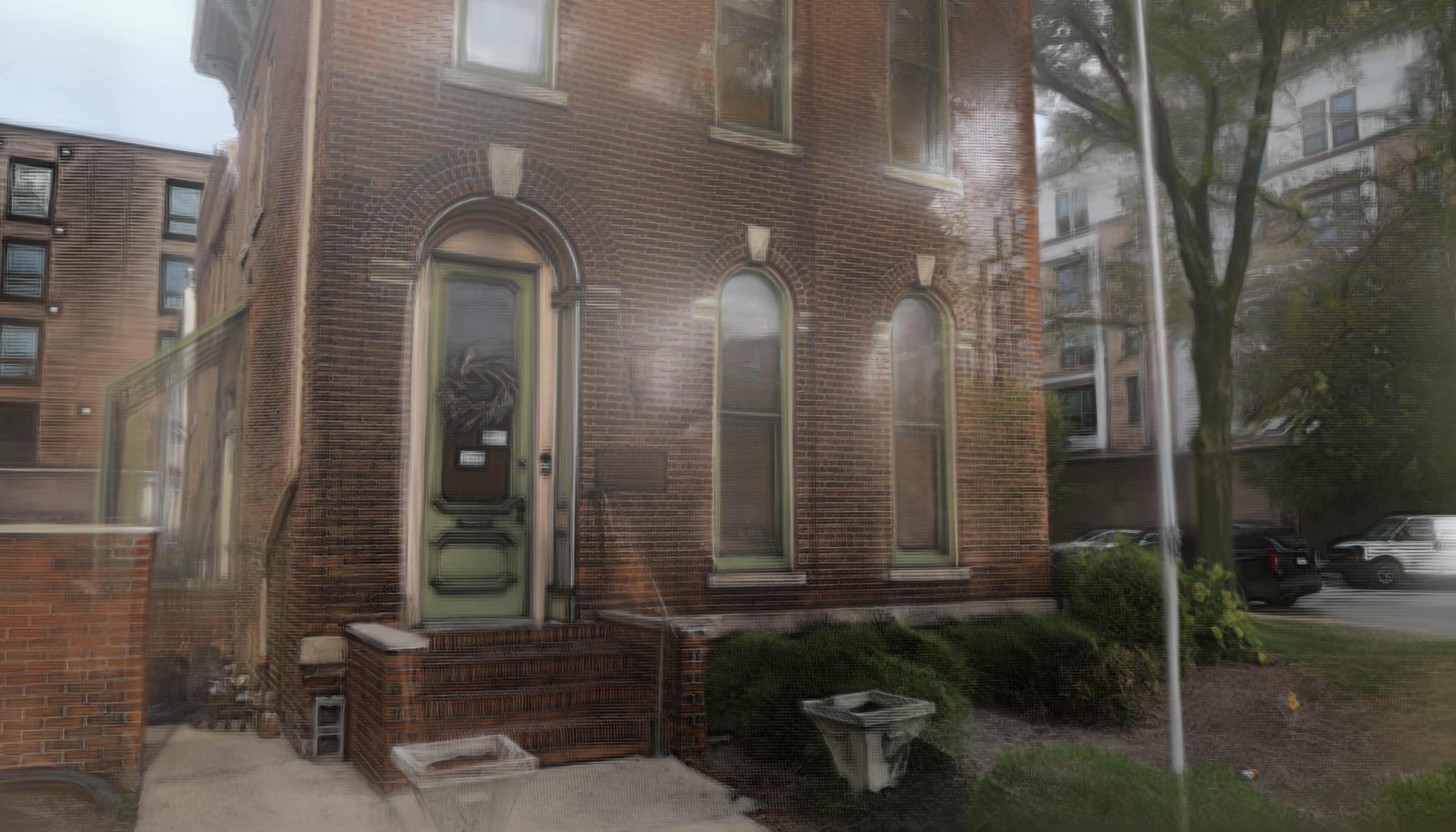} &
\includegraphics[width=\linewidth]{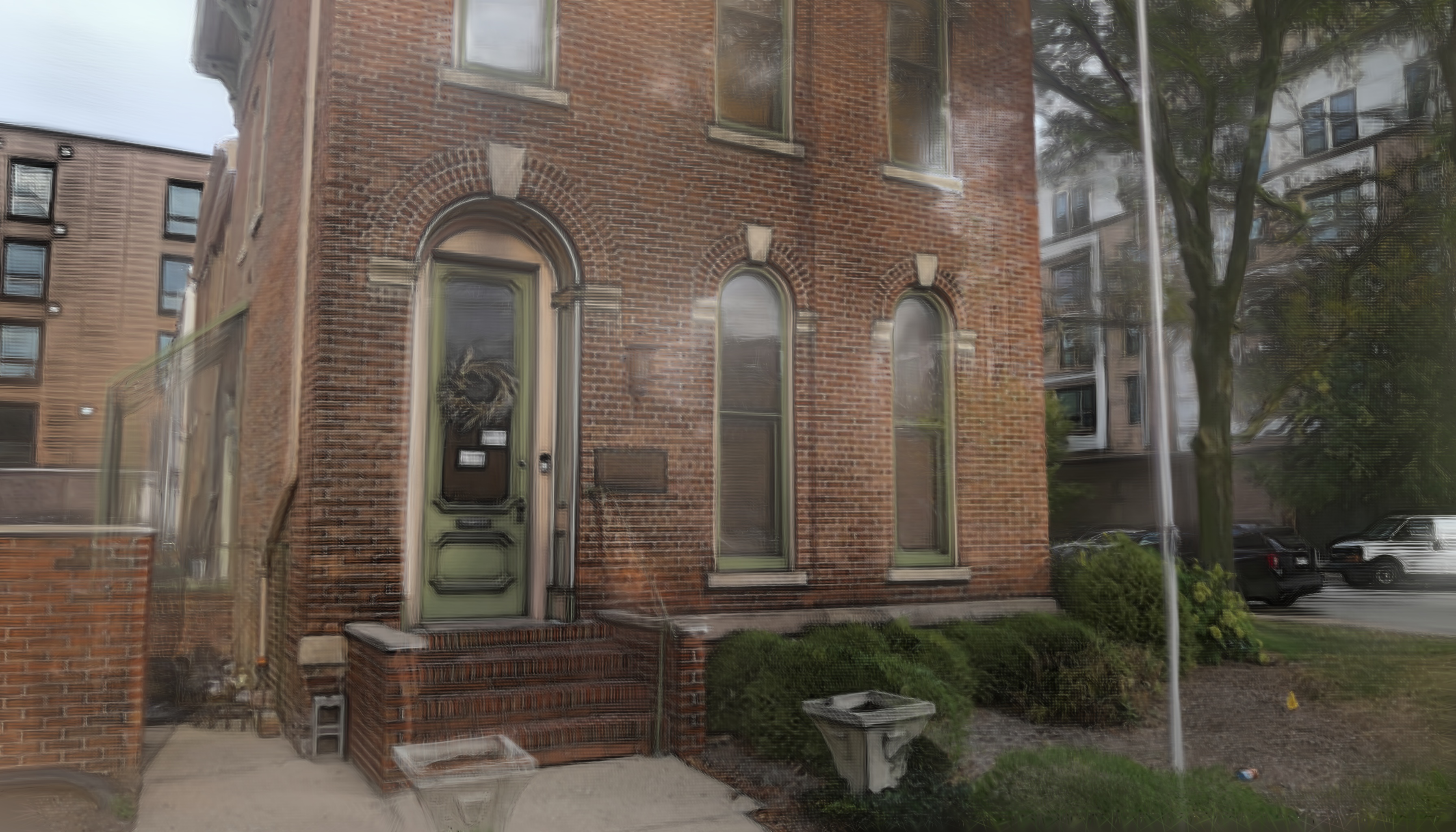} &
\includegraphics[width=\linewidth]{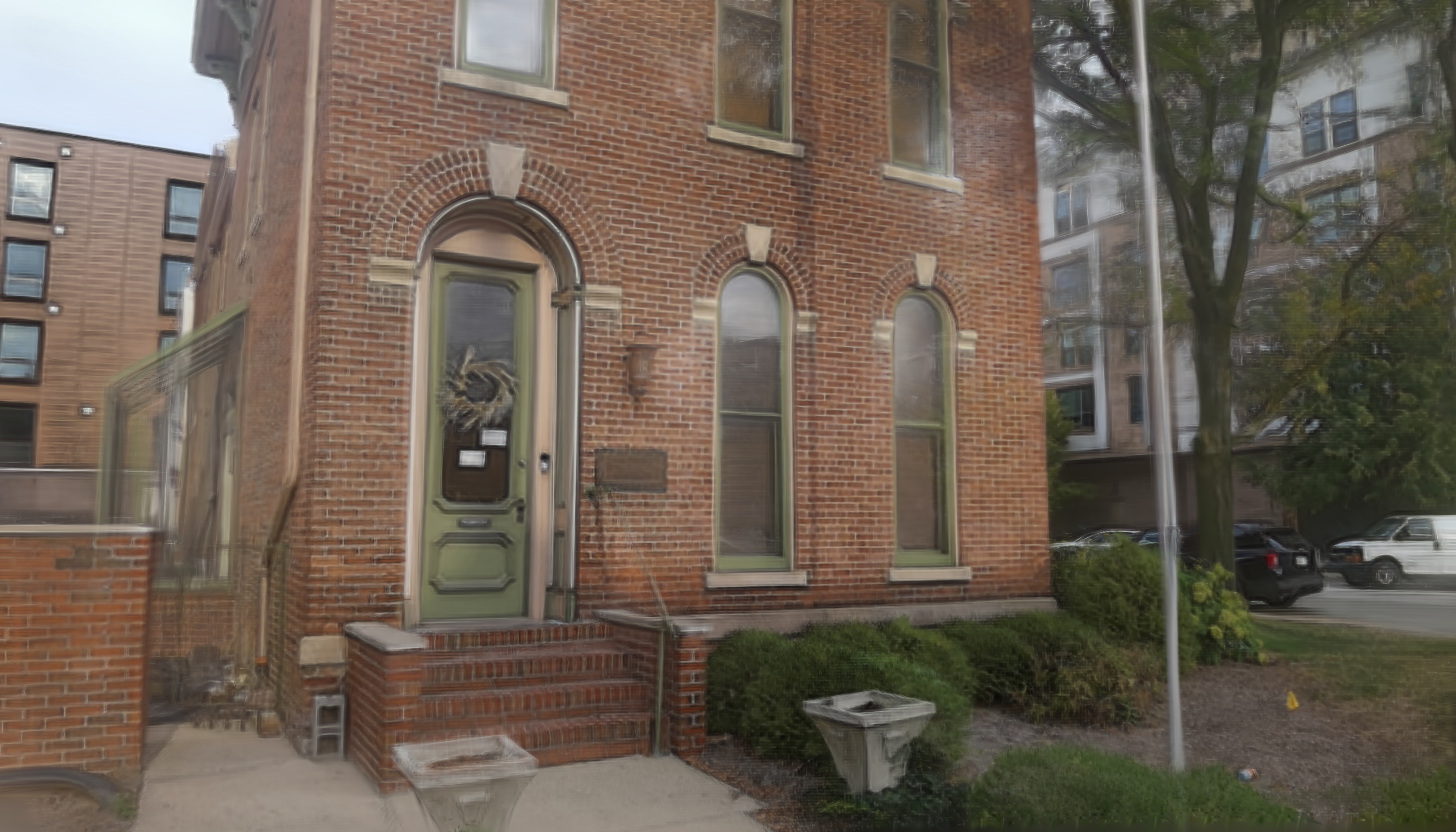}  \\

\end{tblr}
\vspace{-3mm}
\caption{\textbf{Ablation study on DL3DV}. Alpha normalization improves the robustness to varying input views, while 3D sampling-based regularizer improves high-resolution rendering. }
\label{fig:ablation_study}
\end{figure}

In this section, we conduct ablation studies to demonstrate the effectiveness of each proposed module. Tab.~\ref{tab:ablation_study} shows the contributions of alpha normalization and 3D sampling-based regularizer. Without these modules, the model reduces to the original DepthSplat~\cite{xu2025depthsplat}. Incorporating \textbf{alpha normalization} improves the model's robustness to more input views, increasing PSNR by {0.95}~dB in the 24-view setting. Moreover, training with alpha normalization also benefits high-resolution rendering, yielding a PSNR gain of {1.09}~dB. Adding \textbf{3D sampling-based regularizer} enhances high-resolution rendering while maintaining comparable performance in the other settings. In particular, at $2048\times3584$ resolution, combining both modules improves PSNR by {2.36}~dB over the original DepthSplat, and by {1.27}~dB compared to using alpha normalization alone. As a result, the final model with both modules achieves the best overall performance across different settings. Fig.~\ref{fig:ablation_study} shows that DepthSplat exhibits noticeable artifacts when increasing the number of input views and the rendering resolution. Adding alpha normalization improves the results when increase the input views, while the full model further improves the high-resolution rendering quality.

% \textcolor{red}{Add description for figures here.}

% we need a conclusion section.

\vspace{-2mm}
\section{Conclusion}
\vspace{-2mm}
This work introduces two components to improve the robustness of feed-forward Gaussian splatting methods. First, we propose an alpha normalization approach that reweights the Gaussians' contributions based on their overlap counts, thereby addressing over-bright renderings and improving robustness to varying numbers of input views. Second, we present a 3D sampling-based regularizer that encourages the prediction of Gaussians with sufficiently large scales, helping to alleviate the hole artifacts that prior methods exhibit when rendering images at high-resolution. Experimental results show that integrating our modules into prior works consistently yields significant improvements under varying input-view counts and high-resolution rendering.

\noindent\textbf{Limitations.} Alpha normalization relies on depth consistency check which could be challenging in case of noisy depth. Moreover, the proposed regularizer requires more training time than prior works.

% Results. 
% Our method is network agnostic and plug and play. 

% Limitation

% \include{supps}

% ─── Bibliography ──────────────────────────────────────────────────────────────
% \clearpage            % force everything out before the refs
{\small               % make the font a bit smaller, as NeurIPS does
  \bibliographystyle{ieeenat_fullname}
  \bibliography{main}  % points to main.bib
}

\newpage

\end{document}